 \documentclass[final,5p,times,twocolumn]{elsarticletba}
\usepackage[numbers]{natbib}

\usepackage{booktabs} 

\usepackage{amsmath}
\usepackage{amssymb}
\usepackage{latexsym}
\usepackage{paralist}
\usepackage{soul, color}
\sethlcolor{green}
\usepackage{listings}
\usepackage{array}
\newcolumntype{P}[1]{>{\centering\arraybackslash}p{#1}}
\newcolumntype{M}[1]{>{\centering\arraybackslash}m{#1}}
\usepackage{tikz}
\usepackage[flushleft]{threeparttable}
\usepackage{etoolbox}
\AtBeginEnvironment{quote}{\singlespacing\small}
\usepackage{lineno}
\usepackage{enumitem}
\usepackage{amsthm}
\usepackage{url}

\usepackage{etoolbox}
\AtBeginEnvironment{quote}{\singlespacing\small}

\newtheorem{definition}{Definition}
\newtheorem{example}{Example}

\lstdefinelanguage{fss}
{
  basicstyle=\ttfamily\footnotesize,
  morestring=[b][\color{black}]\",
	morekeywords={@prefix,ex,deaf,dag,lime,void,hw,rdf,xsd,ontolex,dbr,dct,rdfs,lexinfo,foaf,olia,decomp,vartrans,owl,skos,mola},
  morecomment=[l]{\#},
  stringstyle=\color{black},
  identifierstyle=\color{black},
  upquote=true
}

\emergencystretch=\maxdimen
\def\tsc#1{\csdef{#1}{\textsc{\lowercase{#1}}\xspace}}
\tsc{WGM}
\tsc{QE}
\tsc{EP}
\tsc{PMS}
\tsc{BEC}
\tsc{DE}

\begin{document}
\let\WriteBookmarks\relax
\def\floatpagepagefraction{1}
\def\textpagefraction{.001}

\title{A Review of Multilingualism in and for Ontologies}

\author[mymainaddress]{Frances Gillis-Webber} 
\cortext[mycorrespondingauthor]{Corresponding author}
\ead{fgilliswebber@cs.uct.ac.za}

\author[mymainaddress]{C. Maria Keet} 
\ead{mkeet@cs.uct.ac.za}

\address[mymainaddress]{Department of Computer Science, University of Cape Town, Cape Town, 7701, South Africa}

\begin{abstract}
The Multilingual Semantic Web has been in focus for over a decade. Multilingualism in Linked Data and RDF has shown substantial adoption, but this is unclear for ontologies since the last review 15 years ago. One of the design goals for OWL was internationalisation, with the aim that an ontology is usable across languages and cultures.
Much research to improve on multilingual ontologies has taken place in the meantime, and presumably  multilingual linked data could use multilingual ontologies. 
Therefore, this review seeks to (i) elucidate and compare the modelling options for multilingual ontologies, (ii) examine extant ontologies for their multilingualism, and (iii) evaluate ontology editors for their ability to manage a multilingual ontology.  

Nine different principal approaches for modelling multilinguality in ontologies were identified, which fall into either of the following approaches: using multilingual labels, linguistic models, or a mapping-based approach. They are compared on design by means of an ad hoc visualisation mode of modelling multilingual information for ontologies, shortcomings, and what issues they aim to solve. For the ontologies, we extracted production-level and accessible ontologies from BioPortal and the LOV repositories, which had, at best, 6.77\% and 15.74\% multilingual ontologies, respectively, where most of them have only partial translations and they all use a labels-based approach only. Based on a set of nine tool requirements for managing multilingual ontologies, the assessment of seven relevant ontology editors showed that there are significant gaps in tooling support, with VocBench 3 nearest of meeting them all. This stock-taking may function as a new baseline and motivate new research directions for multilingual ontologies.
\end{abstract}


\maketitle


\section{Introduction}
\label{sec:introduction}

The Semantic Web was envisaged as an extension of the World Wide Web \cite{BernersLee:Hendler:Lassila:2001}.
To support this vision, several languages were developed, notably Resource Description Framework (RDF) and Web Ontology Language (OWL).
We focus on the latter and in particular its internationalisation goal \cite{SemanticWeb:2015,OwlUseCases:2004}, with multilingualism as a common component in an interconnected world. 

\begin{figure*}[t]
\centering
\includegraphics[width=0.995\textwidth]{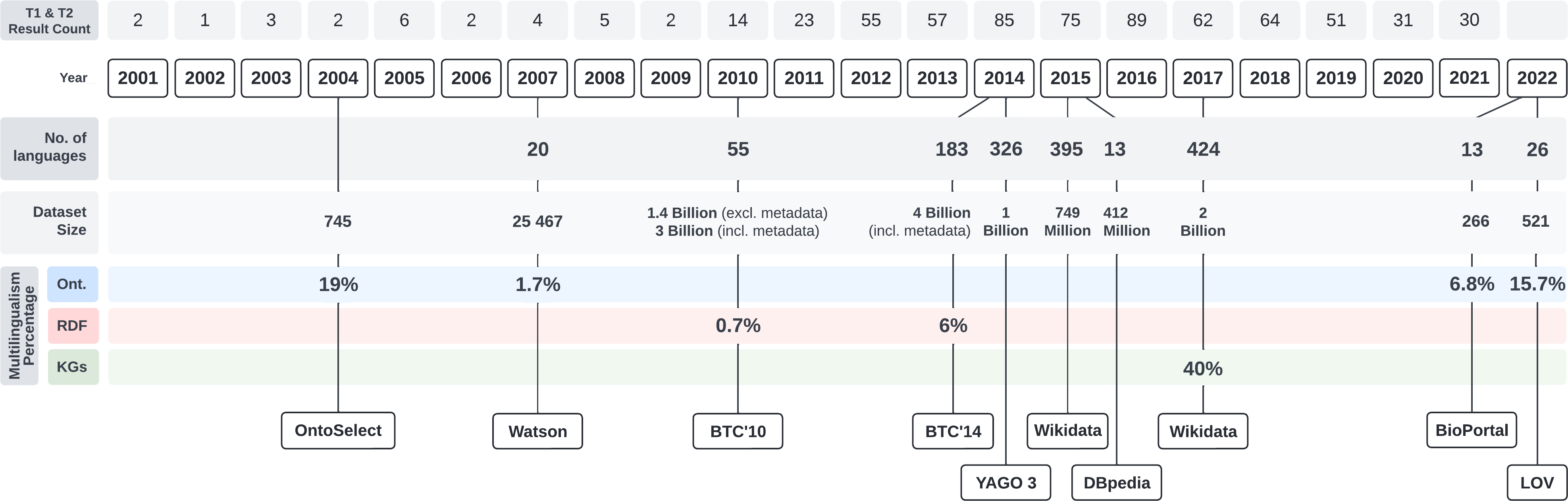}
\caption{Timeline of the MSW (T1) and MWD (T2) searches, contrasted with reviews pertaining to multilingualism conducted for the period 2001--2021: available data for ontologies, RDF datasets (collections of triples) and knowledge graphs (RDF graphs) are shown in the blue, red and green bars respectively.}
\label{fig:timeline}
\end{figure*}

The Semantic Web landscape has grown substantially since the 2000's, and its multilinguality since \textit{circa} 2010 with  the `Multilingual Semantic Web' (MSW) and `Multilingual Web of Data' (MWD) being two widely used terms in the literature \cite{Lemon:2020,Gracia:EtAl:2012,Buitelaar:Cimiano:2014,McCrae:Gracia:2019}.
Indexed research literature illustrate this; e.g., a Google Scholar search on the date range 2001--2005 returned 14 hits for MSW and zero for MWD, for 2006--2010, it increased to 27 hits for MSW and still with zero for MWD, for  2011--2015 it comparatively exploded to 296 for both terms combined, with another 298 for the period 2016--2020. This increase is not as evident for the various stock-takings of multilingual ontologies and related artefacts specifically \cite{Buitelaar:Eigner:Declerck:2004,Daquin:2007,Gillis-Webber:Keet:2022}, as shown in Figure~\ref{fig:timeline}. 
Those 2004 and 2007 reviews on ontologies \cite{Buitelaar:Eigner:Declerck:2004,Daquin:2007} focussed on language-tagged strings, which was practically the only option available in the early years of the Semantic Web, and do not mention the extent of label coverage within an ontology for each natural language, relative to its total class and property axiom count. The very recent review for BioPortal-indexed ontologies \cite{Gillis-Webber:Keet:2022} considered only domain ontologies with `production' status on limited metrics.

This raises several questions, including whether the increase in research on the Multilingual Semantic Web has translated to an increase in the ratio of multilingual ontologies in a repository, compared to the latest review in 2007, as it did for support for natural languages in RDF datasets and knowledge graphs (elaborated on below in Section~\ref{sec:related_work}). With this review we seek to answer the following main questions (motivated afterward):
\begin{compactitem}
\item[RQ1] What are the available modelling options to develop a multilingual ontology? 
\item[RQ2] Regarding extant ontologies and multilingual ontologies: 
\begin{compactitem}
\item[RQ2a] What are the modelling approach(es) used for extant multilingual ontologies?
\item[RQ2b] What is the percentage of multilingual OWL ontologies compared to monolingual ones (in popular repositories)?
\item[RQ2c] What is the percentage of natural language completeness of each multilingual ontology?
\item[RQ2d] How does that compare the the latest stock-taking, of the 2007 review?
\end{compactitem}
\item[RQ3] What is the status of tooling support for developing multilingual ontologies?
\end{compactitem}
First, the aforementioned increase in research papers does include numerous approaches for modelling language information in or for ontologies well beyond a simple language-tagged string, which may benefit from a systematic comparison (RQ1). This is also needed in order to be able to find multilingual ontologies and to assess them (RQ2) on which multilingual approaches are actually used, and how many multilingual ontologies there are. Their presence, or absence, may well be linked to tooling support to be able to conveniently develop and maintain them, and, hence, RQ3.

To answer the questions, the first step was to analyse proposed and related variant approaches to multilingualism in ontologies. 
To facilitate comparison, we devised a visualisation notation for them. 
Nine main approaches were identified, which can be grouped into the categories of multilingual labels, linguistic models, and mapping-based strategies. 
For the assessment of ontologies, we consulted NCBO~BioPortal\footnote{\url{https://bioportal.bioontology.org/ontologies}}, a repository of biomedical ontologies \cite{Salvadores:EtAl:2013}, since there has been wide adoption of (OWL) ontologies in the biomedical domain since the early 2000s \cite{Yu:2006,Hoehndorf:Dumontier:Gkoutos:2013, Hoehndorf:Schofield:Gkoutos:2015}, and Linked Open Vocabularies (LOV)\footnote{\url{https://lov.linkeddata.es/dataset/lov/}}, a curated repository of RDF and OWL vocabularies that is not limited to any one specific domain. 
Further, we looked beyond the mere number of multilingual ontologies, to also assess how multilingual the more-than-one-language ontologies are, introducing the notion of coverage within an ontology and language-specific completeness. 
Only a few of the multilingual modelling methods are used, principally just in the labels category, and even less can be considered fully multilingual where there is more than one primary language. 
Tooling for developing and managing multilingual models is also very limited, where of the several relevant ones, none meets all the requirements specified.

In the remainder of this paper, we first describe related work and key reviews of RDF datasets and knowledge graphs  in Section~\ref{sec:related_work}. 
The ways to model multilinguality is described and compared in Section \ref{sec:modelling_multilinguality}. 
This is followed by the BioPortal and LOV evaluations in Section~\ref{sec:review_ontologies_bioportal}, and the tools assessment in Section~\ref{sec:managing_multilinguality}. 
A discussion of the reviews and specific answers to the research questions is provided in Section~\ref{sec:discussion}. 
The paper concludes with Section \ref{sec:conclusion}.

\section{Related Work}
\label{sec:related_work}

As noted in the Introduction, the sampling of counts of multilingual ontologies has been sparse, with just \cite{Buitelaar:Eigner:Declerck:2004,Daquin:2007,Gillis-Webber:Keet:2022}. We therefore cast the net wider to also include 
RDF datasets and knowledge graph literature pertaining to multilingualism. The ones with such data are included also in Figure \ref{fig:timeline} and they are briefly discussed here. 

Ell et al. \cite{Ell:Vrandecic:Simperl:2011} conducted a review on the Billion Triple Challenge (BTC) 2010 corpus, which  contained over 3~billion triples, including metadata on the source of the resource the triple was crawled from. 
Excluding metadata, 1.4~billion distinct triples remained and of these triples, 0.7\% were identified to contain two or more language tags \cite{Ell:Vrandecic:Simperl:2011}. 
This was followed-up in 2018 with a cross-sectional study of labels across seven datasets \cite{Kaffee:Simperl:2018}, including, among others, the BTC~2010 corpus (with the metadata), a 2014 version of BTC (4~billion triples), and a 2017 version of Wikidata (2~billion triples). 
BTC~2014 was found to support 183 languages (up from 2010's 55) and the Wikidata dataset was found to support 424 languages \cite{Kaffee:Simperl:2018}. 
That is: languages other than English certainly are being used in the Semantic Web. 
Due to the way the results were reported, however, it is not possible to determine an accurate percentage of the number of triples containing two or more multilingual labels, but it is an approximate 6\% for BTC~2014 and 40\% for Wikidata\footnote{The data was reported in groups of 2, 2--5, 5--10, and $>$10. 2 was ignored as it is included in 2--5. The sum of the groups for BTC~2014 was 6.3\% and for Wikidata, 4.4\%. Both values were rounded down due to the unavoidable double-counting of 5.}. Coverage was measured for languages together, rather than by language.

\begin{table*}[t]
\begin{threeparttable}
\fontsize{9}{12}\selectfont
\caption{\label{table:overview_multilingual_options} Comparison of the Main Approaches for Modelling Multilinguality in Ontologies}
\begin{tabular}{ | p{0.446\linewidth} | p{0.17\linewidth} | p{0.14\linewidth} | p{0.13\linewidth} |}
\hline
 & \textbf{Labels} & \textbf{Linguistic Models} \tnote{\dag} & \textbf{Mapping Models} \\
\hline \hline
Possible OWL profiles & All & All & All \\ 
\hline
Can account for cross-lingual terms where there is no 1-1 mapping & No & Within the annotation only & Yes \\ 
\hline
Can support inflectional languages & No & Limited & No \\
\hline
Annotations reusable by other resources & $\pm$ (not if the annotation is a data literal) & Yes & No \\ 
\hline
Example axiom count for two natural languages in an ontology \tnote{\ddag} & 2 & $\geq$14 & 7 \\ 
\hline
Annotations contained within the OWL file & Yes & No & No \\ 
\hline
Number of files to keep in sync & 1 & Minimum 3 & Minimum 2 \\ 
\hline
Can be managed by an ontology editor & Yes & Limited support & Limited support \\ 
\hline
\end{tabular}
\begin{tablenotes}
\item[\dag] Values are for OntoLex-Lemon only for the inflectional languages, axiom count, and ontology editor.
\item[\ddag] The axiom count was determined for a class in an ontology for two natural languages. Language-specific grammatical features were excluded. OntoLex-Lemon was used for the linguistic model and the simplest method was used, that is, associating two lexical entries with a class. For the mapping model, OWL~2 was used.
\end{tablenotes}
\end{threeparttable}
\end{table*}

Other dataset and knowledge graph assessments also indicate substantial use of multiple languages. 
Notably, the 2015-10 version of Wikidata was found to support 395 languages in 749~million triples, the 2015-04 version of DBpedia supported 13 languages with 412~million triples, and Yet Another Great Ontology (YAGO), which integrates statements of the different language versions of Wikipedia, and YAGO~3 (with the 2014 Wikipedia dump) was found to support 326 languages in 1~billion triples \cite{Farber:EtAl:2018}.
Completeness was discussed as a data quality dimension on whether the knowledge graph was suited to the task at hand, within the context of its data. The coverage of selected languages (English, French, German, Spanish, and Italian) was considered relative to the annotations: German and French had a coverage of over 30\% in Wikidata, and German of over 10\% in YAGO.

Given the increase in data in multiple languages and an increase in multilingualism over the years, one may expect an increase in multilingual ontologies as well.

\section{Modelling Options for Multilingual Ontologies}
\label{sec:modelling_multilinguality}

Several options to develop multilingual ontologies have been identified \cite{Gracia:EtAl:2012,MontielPonsoda3:EtAl:2011,Adamou:EtAl:2012}:
\begin{enumerate}
\item {\em Multilingual labels}: the ontology vocabulary is annotated with language-tagged strings.
\item {\em Linguistic models}: the ontology is associated with a linguistic model external to it.
\item {\em Mapping models}: common interlingua which can be specified between one or more monolingual ontologies.
\end{enumerate}
A brief comparison of these approaches is included in Table \ref{table:overview_multilingual_options}. We will elaborate, compare, and discuss each approach in turn in the subsections below.

\begin{figure}[t]
    \centering
    \includegraphics[width=0.97\linewidth]{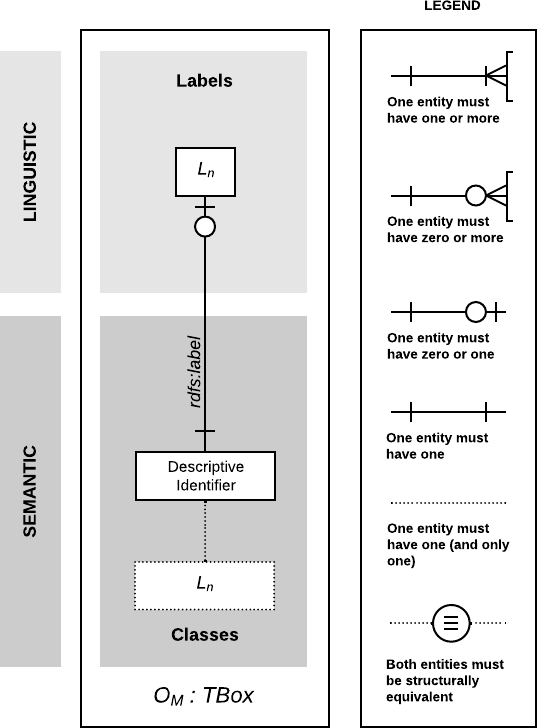}
    \caption{Demonstration of the language to visualise the interaction between ontologies, any identifiers, and natural language aspects, shown here for a monolingual ontology that uses descriptive identifiers. The semantic and linguistic layers of the ontology are indicated by the grey boxes; the names of the classes (and other vocabulary elements) are in a natural language $L_n$. The possible relations with their constraints are shown in the legend.}
    \label{fig:legend}
\end{figure}

In order to represent the options and sub-variants schematically to facilitate comparisons, we use an ad hoc visual language. This consists of a set of primitives. First, an ontology can be visualised as two layers, which separate the semantic and the linguistic components, in line with \cite{Cimiano:EtAl:2010}. The semantic layer, which contains the classes, object properties and other axioms, is indicated with a dark grey box that has further information, such as the type of identifier (opaque or descriptive). 
The linguistic layer, which typically contains the annotations used to associate language information with an ontology, is indicated by a light grey box, and contains the various options for the language annotations.
An example visualisation is shown in Figure~\ref{fig:legend} for a monolingual ontology with descriptive identifiers, i.e., naming the vocabulary elements with a string that is in a natural language or natural language-like (indicated with $L_n$) and meaningful to a human, such as naming a class {\sf Person} or {\sf RumAndRaisinIcecream} instead of {\sf ABC:0000012}. 
The knowledge represented in the TBox of the ontology may have a natural language-specific vocabulary either in or attached to the ontology as a separate file. It is possible that two monolingual ontologies in two different natural languages may share the same conceptualisation \cite{Guarino:1998}.

To show the properties and cardinality constraints between the two layers and their objects, the crow's feet notation from Entity-Relationship modelling is used.
The notation is extended to indicate the modelling of language-specific label instances.
We also introduce a relation that is used to indicate when two elements must have structural equivalence.
This pertains specifically to the different annotation values available in OWL~2.
See Figure~\ref{fig:legend} for a legend of the different relations.

We now proceed to the three groups of options, going from the basic to more elaborate options.  

\begin{figure*}[t]
\centering
\includegraphics[width=0.7\textwidth]{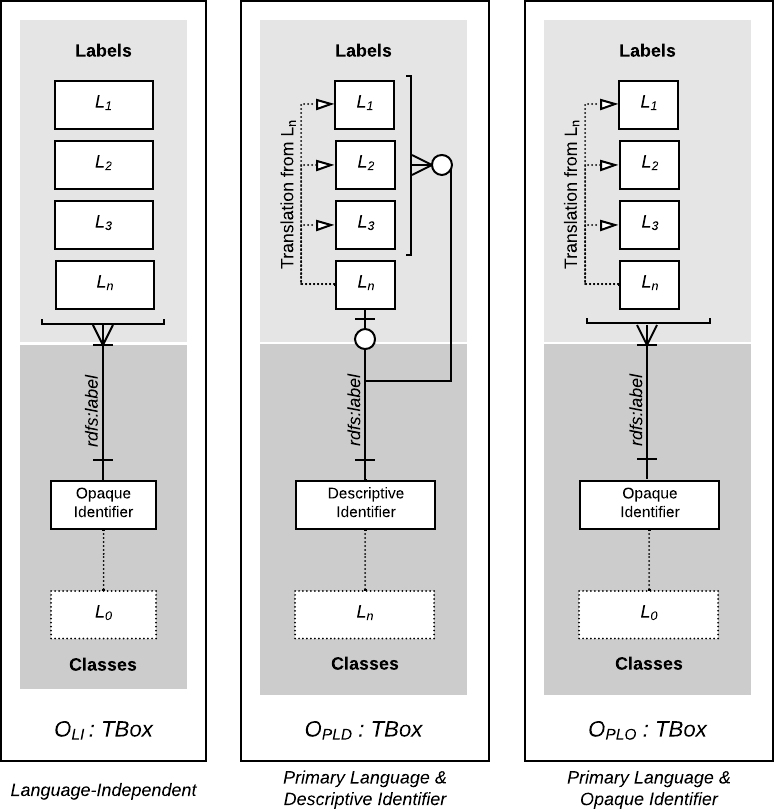}
\caption{Three principal options for multilingual labels in the TBox of the ontology: fully language independent (indicated with $L_0$) with a mandatory requirement for at least one label in some language (any one of $L_1 \ldots L_n$) (left), a primary language for the ontology ($L_n$), using a descriptive identifier and optional labels for other languages (centre), and mostly language-independent ($L_0$ in the semantic later) but with the requirement to have at least one label in one specified language (right)}
\label{fig:model_languageIndependent}
\end{figure*}

\subsection{Multilingual Labels}
\label{sec:modelling_multilinguality_labels}

Multilingual labels are the `low-hanging fruit', being the easiest way to develop a multilingual ontology.
An entity in an ontology (i.e., a class, object property, data property, or individual) can have multilingual labels by using  
{\tt rdfs:label}, which takes as domain {\tt rdfs:resource} and as range an {\tt rdfs:Literal}.
Each such label is language-tagged.
The language tag typically consists of an ISO 639 language code\footnote{\url{https://www.iso.org/iso-639-language-codes.html}}.
However, as ISO 639's Parts 1-3 \cite{ISO:nd} do not provide language codes for all the world's languages, the language tag can be extended to capture other information, as long as it conforms to IETF's BCP 47 \cite{Phillips:Davis:2006, Phillips:Davis:2009}. When this is insufficient \cite{Gillis:Tittel:2019}, other language models, such as MoLA \cite{Gillis-Webber:Tittel:Keet:2019}, may be added to specify additional languages or lects.

An example OWL fragment of this labelling approach is shown in Listing \ref{fss:multilingual_label} in Functional-Style syntax (FSS).
(Unless explicitly stated otherwise, all further examples are in FSS.)
In this example, the class \textsf{Person} has two annotations: one in English and the other in Dutch.
Both annotations are a language-tagged string.

\begin{lstlisting}[language={fss},caption={OWL fragment illustrating multilingual labels},label=fss:multilingual_label,captionpos=b]
SubClassOf(:Person owl:Thing)
AnnotationAssertion(rdfs:label :Person "Person"@en)
AnnotationAssertion(rdfs:label :Person "Persoon"@nl)
\end{lstlisting}

The class {\tt :Person} is a named entity, where {\tt Person} is the identifier portion of the URI.
An identifier of a URI can be {\em opaque} or {\em descriptive} \cite{Archer:Goedertier:Loutas:2012, LabraGayo:Kontokostas:Auer:2013, MontielPonsoda2:EtAl:2011}.
An opaque identifier is semantic-free (non-meaningful) and used in a number of ontologies; e.g., \cite{Archer:Goedertier:Loutas:2012,Manaf:Bechhofer:Stevens:2010,McMurry:EtAl:2017}. If an opaque identifier is used as part of the URI, then an additional sign in the form of an {\tt rdfs:label} annotation should be added for both humans and machines to interpret the URI. 
For a descriptive identifier, there is a direct relationship between the natural language term used as an identifier and its semantics, e.g., \cite{Manaf:Bechhofer:Stevens:2010,Third:2012}. 
The descriptive identifier is typically in the primary language of the ontology.

Figure~\ref{fig:model_languageIndependent} shows a schematic representation for the TBox for three ways of realising multilingual ontologies with this approach. 
Consider the TBox of \(O\!_{LI} \), which presents the sub-variant where a class has an opaque identifier and at least one natural language label, with each label a different language.
An example OWL fragment is given in Listing \ref{fss:rdfs_labels_owl}. 

\begin{lstlisting}[language={fss},caption={OWL fragment with an opaque identifier},label=fss:rdfs_labels_owl,captionpos=b]
SubClassOf( :493Dk owl:Thing )
AnnotationAssertion( rdfs:label :493Dk "Person"@en )
\end{lstlisting}

\noindent An entity can have any number of labels, where \textit{i} is a label and L is the set of languages used for an entity: (L\textsubscript{1} ... L\textsubscript{n}) with \(1 \leq i \leq n\).
For a primary language ontology, an entity can either have an opaque or descriptive identifier in L\textsubscript{i}.

Because meaning can be surmised by a human from a descriptive identifier, a label is not required.
However, as shown in \(O\!_{P\!L\!D} \) in Figure~\ref{fig:model_languageIndependent}, if the ontology is multilingual, all entities may have labels in one language, with the language tag sometimes omitted.
Where there are labels for other languages, these are translations from the primary language.
\(O\!_{P\!L\!O} \) (Figure \ref{fig:model_languageIndependent}) has an opaque identifier, but since all entities have labels in one language and to a limited extent, labels for other languages, we consider it to be a primary language multilingual ontology. 

The adaptation of a monolingual ontology to another language is called ontology localisation \cite{Surez-Figueroa:Gomez-Perez:2008}, although this is used for different contexts as well, such as a culture or a geo-political environment for which there is typically a shared language \cite{Cimiano:EtAl2:2010, Bouquet:EtAl:2004, Leon-Arauz:Faber:2014}.
If an ontology is localised using labels, then only the linguistic layer of the ontology is affected, and the concept space of the original ontology remains unchanged.

Despite the differences with the labels and identifiers between each of the TBoxes in Figure \ref{fig:model_languageIndependent}, \(O\!_{LI} \), \(O\!_{P\!L\!D} \) and \(O\!_{P\!L\!O} \) have commonality in that they share the same so-called `concept spaces', which can be seen as a figurative area of a concept, from a fine-grained concept to a category of concepts, where a concept space applies to each OWL class and object property; e.g., the notion of house and the colour blue. 
Natural languages may divide up the same concept space differently, which may indicate underlying ontological distinctions. 
One language may underspecify it compared to that of another. For instance, there is only one notion of `River' in English, whereas the French equivalent distinguishes between two types of river (\textit{Rivi\`ere} and \textit{Fleuve}). 
Another example is the representation of part-whole relations in isiZulu when compared to English, which also revealed ontological distinctions, in that the list of `universal' part-whole relations required both generalisation and refinements in order to accurately represent them \cite{Keet:Khumalo:2018}.

Thus the use of labels has its limitations. In addition, not all grammatical features for a natural language can be dealt with by a label,  such as inflected forms, concordial agreement, and agglutination (where words are formed with the addition of affixes to a word root or stem). 
Inflectional forms on nouns include the grammatical categories number and gender, and tense, aspect and number on verbs, which are typically used for naming classes and properties, respectively.
For each of these grammatical categories, the word formation may need to change.
An example is the Organization Ontology \cite{OrgOntology:2014}, which provides support for English, Spanish, French, Italian and Japanese.
For instance, the property \textit{org:changedBy} with the English label \textit{changed by} has labels for most of the supported languages, but there are two for Spanish: \textit{es modificada por} and \textit{es modificado por}.
This is to account for grammatical gender which is determined by the gender of the noun of the name of the class that is in the position of the domain in the axiom.
Similarly for plural terms, the concordial agreement in a language such as isiXhosa (a Niger-Congo~B (`Bantu') language spoken in South Africa), combined with its agglutination, will alter the surface realisation of the annotation depending on its domain and range.
Indeed, as highlighted by Keet and Khumalo, highly agglutinative languages can represent a challenge in ontologies, where it is possible that no single human-readable label can be prescribed to a property, due to the use of context-dependent affixes that modify the entity's name or their label \cite{Keet:Khumalo:2018}. 

\begin{table*}[t]
\fontsize{9}{12}\selectfont
\caption{\label{table:overview_multilingual_labels} Overview of the Limitations of Multilingual Labels}
\begin{tabular}{ | p{0.3\linewidth} | p{0.65\linewidth} |}
\hline
\textbf{Feature/Approach} & \textbf{Shortcoming/Limitation} \\
\hline \hline
Natural language identification & The language tag appended to a label identifies the natural language. If this language tagging is not done, or incorrectly, or a custom language tag is created, then human inspection is required to identify the language or a tool needs to be equipped to parse the language tag. \\ 
\hline
Ease of use for realising multilingualism & Although multilingual labels are easy to add, this applies to the linguistic layer of the ontology only. Where there is no 1-1 mapping of the terms for an underlying conceptualisation, the target term will be an approximation of the primary language, losing precision that an ontology is supposed to provide. \\ 
\hline
Handling of certain grammatical features such as gender, agglutination, concordial agreement, and other highly inflectional forms & When these features result in word formation changes, it is not possible to specify these differing word formations in a single label. \\
\hline
Use of non-standard labels & When custom annotation properties are created for the purpose of labelling, human inspection is required to identify these properties as labels. This is problematic when ontologies are used on-the-fly by an application. \\  
\hline
\end{tabular}
\end{table*}

It may be the case that more than one label for a single language has to be added, fo which Simple Knowledge Organization System (SKOS) can be used. SKOS has the {\tt prefLabel} and {\tt altLabel} properties (subsumed by {\tt rdfs:label}) where labels can be identified as preferred or alternative,  respectively. {\tt rdfs:label} does not have cardinality restrictions, but {\tt skos:prefLabel} can only be used once per natural language for an entity.

For a succinct list of limitations of multilingual labels, see Table \ref{table:overview_multilingual_labels}.

\subsection{Linguistic Models}
\label{sec:modelling_multilinguality_linguistic}

An ontology can be associated with an external linguistic model, where the lexicon is modelled separately from the ontology. 
A lexicon is a collection of lexical entries and each lexicon is language-specific \cite{Cimiano:Mccrae:Buitelaar:2016}, which, within the context of the Semantic Web, is typically stored in any one of the RDF serialisations. 
This approach is in contrast to the aforementioned multilingual labels approach that has all linguistic annotations in the same OWL file. To use a linguistic model in practice, one needs a linguistic model, a way to store the language data must be devised, as well as a way to connect the resources.

Many linguistic models have been developed over the years, such as LingInfo \cite{Buitelaar:EtAl:2006}, LexOnto \cite{Cimiano:EtAl:2007}, Linguistic Information Repository (LIR) \cite{Peters:EtAl:2007, MontielPonsoda:EtAl:2008}, Linguistic Watermark 3.0 \cite{Pazienza:Stellato:Turbati:2008}, LexInfo \cite{Cimiano:EtAl:2010, Buitelaar:EtAl:2009}, \textit{lemon} \cite{Lemon:2020, McCrae:EtAl:2012}, and OntoLex-Lemon \cite{Cimiano:Mccrae:Buitelaar:2016, Mccrae:EtAl:2017} (based on \textit{lemon}).
As OntoLex-Lemon is now the \textit{de facto} standard for describing linguistic resources in RDF \cite{Cimiano:Mccrae:Buitelaar:2016}, this model is the only one considered for comparative analysis.

\begin{figure}[h]
    \centering
    \includegraphics[width=1.0\linewidth]{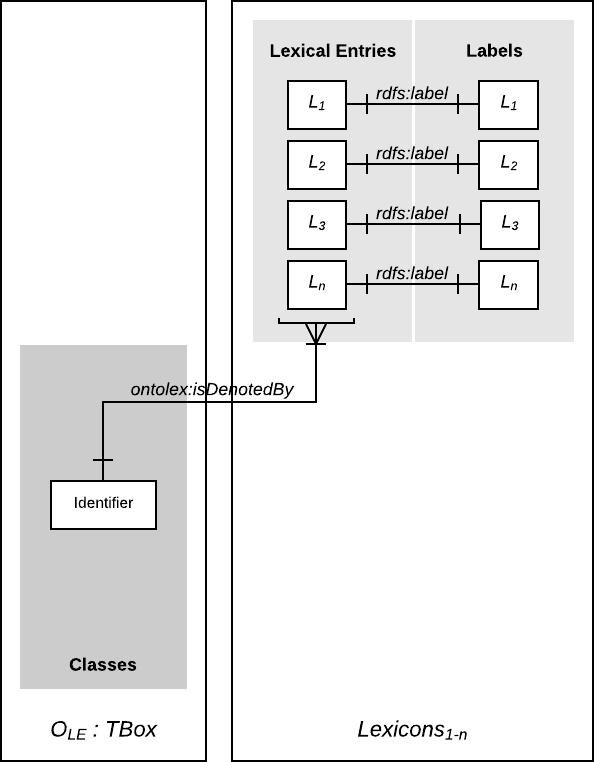}
    \caption{Visualisation of the use of the OntoLex-Lemon linguistic model, where lexical entries are linked to an entity in the ontology (shown here for a class). The identifier of the entity can be either language-independent ($L_0$) or any of the natural languages and therefore not show in the dark-grey semantics box on the left.}
    \label{fig:model_linguisticEntries}
\end{figure}

The ontology-lexicon interface is done in OntoLex-Lemon using the {\em semantics by reference} principle, whereby the semantics of a lexical entry (be it a word, affix or multi-word expression) is expressed by an entity in an ontology \cite{Buitelaar:2010}. 
This entity can be a class, property or individual. 
The ontology-lexicon interface can be modelled in several ways, each with increasing degrees of complexity.
In Figure \ref{fig:model_linguisticEntries}, an element in the ontology is linked to one or more lexical entries, with each lexical entry in a different language.
An example OWL fragment for a class is given in Listing \ref{fss:linguistic_model_lexical_entry_owl}, with a corresponding lexical entry in RDF given in Listing~\ref{fss:linguistic_model_lexical_entry_rdf}.

\begin{lstlisting}[language={fss},caption={OWL fragment linking a lexical entry to an entity},label=fss:linguistic_model_lexical_entry_owl,captionpos=b]
SubClassOf( :493Dk owl:Thing)
AnnotationAssertion(rdfs:label :493Dk "Person"@en)
AnnotationAssertion(ontolex:isDenotedBy :493Dk
                     ex:en/lexicalEntry_Person)
AnnotationAssertion(ontolex:isDenotedBy :493Dk
                     ex:nl/lexicalEntry_Persoon)
\end{lstlisting}

\begin{lstlisting}[language={fss},caption={RDF Turtle fragment of a lexical entry that is stored in a file separate from the ontology.},label=fss:linguistic_model_lexical_entry_rdf,captionpos=b]
:nl/lexicalEntry_Persoon
    a ontolex:LexicalEntry ;
    dcterms:language lang:dutch ;
    rdfs:label "Persoon"@nl ;
    ontolex:canonicalForm  
       :nl/lexicalEntry_form_Persoon .

:nl/lexicalEntry_form_Persoon
    a ontolex:Form ;
    ontolex:writtenRep "persoon"@nl .
\end{lstlisting}

\begin{figure*}[t]
    \centering
    \includegraphics[width=0.7\linewidth]{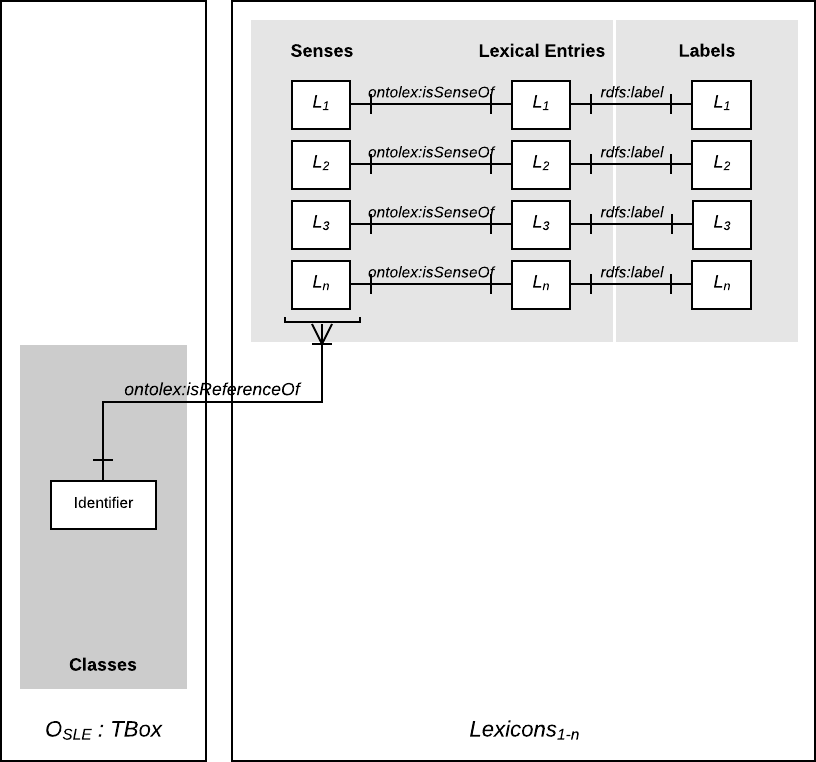}
    \caption{Visualisation of the OntoLex-Lemon linguistic model in a more elaborate setting cf. Figure~\ref{fig:model_linguisticEntries}: lexical senses are linked to an entity in the ontology and each lexical sense is related to a lexical entry.}
    \label{fig:model_linguisticSenses}
\end{figure*}

As a lexical entry can have different meanings depending on the context, it can have more than denotation.
This introduces ambiguity; to resolve this difference in meaning, senses can be modelled within a lexical entry, where the sense is then linked to the entity in the ontology \cite{Cimiano:Mccrae:Buitelaar:2016}, as shown in Figure~\ref{fig:model_linguisticSenses}. 
A lexical sense can only be linked to one lexical entry, and it can also only have one reference in an ontology.

The modelling of an inflected form such as grammatical gender is possible in OntoLex-Lemon.
If a term is distinct in gender or inflected forms, then a separate lexical entry is created for each form.
For a noun, if the gender holds for all its forms, then one lexical entry suffices. 

Revisiting the \textit{changed by} property from Section \ref{sec:modelling_multilinguality_labels}, six lexical entries are required for Spanish: a compound lexical entry each for the feminine and masculine using the {\tt decomp:constituent} predicate from OntoLex-Lemon, and four for the individual words.
In contrast, only three lexical entries are required for English: one for the present infinitive \textit{`change'}, another for \textit{`by'}, and \textit{`changed by'} is then modelled as a compound lexical entry.
Highly inflectional languages such as isiZulu (a Niger-Congo B language spoken in South Africa), and Chichewa (spoken in Southern, South-East, and Eastern Africa) could not be modelled in OntoLex-Lemon's more expressive predecessor \textit{lemon} \cite{Chavula:Keet:2014}. 
The Morphology module for OntoLex-Lemon remains in development, and perhaps in the future it might become possible.

Revisiting the notion of the concept space that was introduced in Section \ref{sec:modelling_multilinguality_labels}, OntoLex-Lemon allows for ``meaning nuances'' of a word to be modelled using the {\tt ontolex:usage} property \cite{Cimiano:Mccrae:Buitelaar:2016}.
Continuing with the example of the English `River' to its French equivalents {\em Rivi\`ere} and {\em Fleuve}, the same example is given in \cite{Cimiano:Mccrae:Buitelaar:2016} to demonstrate the {\tt ontolex:usage} property.
The usage condition is given by the definition of {\em Fleuve} in English; however, identification of when to apply this usage condition requires human inspection, therefore {\tt ontolex:usage} is not ideal for automated processing. 
If the reference or denotation can be made more specific in the ontology, then a possible solution is to represent {\em Rivi\`ere} and {\em Fleuve} as sub-classes of {\sf River}, with an associated {\sf flows into} object property; e.g.,:

\({\sf Riviere \sqsubseteq River \sqcap \exists flowsInto.(Riviere \sqcup Fleuve) }\) \newline
\indent \({\sf Fleuve \sqsubseteq River \sqcap \exists flowsInto.Sea} \) \newline
\indent \({\sf Riviere \sqsubseteq \neg Fleuve} \)

\noindent Linguistic models that rely on a domain ontology for providing the sense inventory are criticised by Hirst \cite{Hirst:2014}.
According to Hirst, there simply is insufficient granularity provided by a domain ontology, particularly once multilinguality is factored in, due to the different sets of word senses across languages~\cite{Hirst:2014}.

A challenge with a linguistic model such as OntoLex-Lemon is that the lexical entries for each natural language will differ in terms of how it uses the linguistic model. For instance, English has limited inflection, Spanish has grammatical gender, and isiXhosa has extensive concordial agreement and agglutination. 
Practically, for automatic processing, such as real-time ontology verbalisation, then the application will have to know the peculiarities (and possibly also additional grammar rules) of each natural language through something like a language specific dispatch mechanism.

\begin{table*}[t]
\fontsize{9}{12}\selectfont
\caption{\label{table:overview_linguistic_model} Overview of the Limitations of OntoLex-Lemon as a Linguistic Model for Modelling Multilingualism}
\begin{tabular}{ | p{0.3\linewidth} | p{0.65\linewidth} |}
\hline
\textbf{Feature/Approach} & \textbf{Shortcoming / Limitation} \\
\hline \hline
Use of the semantics by reference principle for meaning & Problematic if the ontology entity is not granular enough or lexical entries with overlapping meanings cannot be accurately referenced or not represented in sufficient detail due to logic limitations. \\
\hline 
To address the granularity/overlapping meanings issue, OntoLex-Lemon proposes the `usage' property & This property is sufficient for displaying hard-coded text as lexical definitions, but when used in real-time by an application that is, for example, verbalising an ontology, the property still requires human inspection. \\ 
\hline
Modelling grammatical gender and similar inflections & Although modelling is possible, the application that parses this data for purposes of display or  verbalisation has to know the grammar rules and have verbalisation templates specific to the language, neither of which can be included in the linguistic model. \\ 
\hline
Modelling highly inflectional languages & Languages such as isiZulu and isiXhosa that have features such as agglutination and concordial agreement, cannot be modelled in OntoLex-Lemon. \\ 
\hline
Modelling of object/data properties & This can be time-consuming as a property may be represented by a compound lexical entry, which in turn consists of multiple lexical entries.  \\ 
\hline
\end{tabular}
\end{table*}

In closing, for an overview of the limitations of OntoLex-Lemon, see Table \ref{table:overview_linguistic_model}.

\subsection{Mapping Models}
\label{sec:modelling_multilinguality_mapping}

Two or more ontologies in different natural languages with overlapping content can be aligned to create a merged ontology, resulting in labels in at least two natural languages.
If there is a common natural language between the ontologies being aligned, then the alignment process can be done on the shared language.
For example, if O\textsubscript{1} is multilingual with languages L\textsubscript{1} and L\textsubscript{2}, and O\textsubscript{2} has languages L\textsubscript{1} and L\textsubscript{3}, then multilingual alignment can be done between both L\textsubscript{1}'s.
If two or more ontologies are aligned and there is no shared natural language between them, then this is referred to as cross-lingual alignment \cite{Trojahn:EtAl:2014}.

Alignment can be done in the semantic or the linguistic layer.
The former is generally 
done between the same type of element, where alignments between classes and between properties are expressed as bridge rules and alignments between individuals are expressed as individual correspondences \cite{Inants:Euzenat:2015}.
Linguistic mappings are declared in the linguistic layer, using, for example, a linguistic model or synsets from WordNet. 
We will discuss the merits of each in turn.

\subsubsection{Mappings in the ontology.}
\label{sec:modelling_multilinguality_mappingConceptual}

\begin{figure}[t]
\centering
\includegraphics[width=0.48\textwidth]{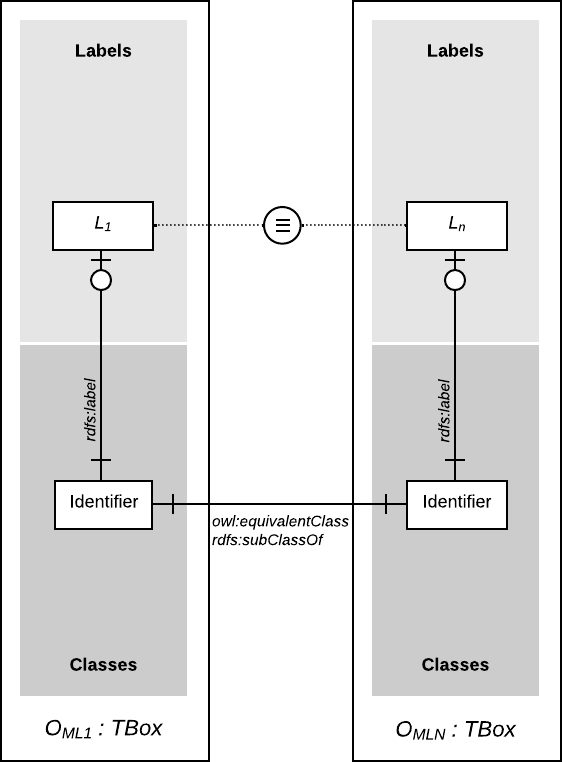}
\caption{Using OWL as a `mapping model' in the TBox: identifiers from monolingual ontologies are mapped to create a multilingual ontology; the ontologies have to be structurally equivalent in OWL.}
\label{fig:model_mappingModelsOwl}
\end{figure}

The ontology alignment relations applicable to classes and properties are equivalence with {\tt owl:EquivalentClass} or subsumption with {\tt rdfs:subClassOf}, as shown in Figure~\ref{fig:model_mappingModelsOwl} and an example for more than two ontologies is shown in Listing \ref{fss:mapping_model_class_equivalence}.

\begin{lstlisting}[language={fss},caption={OWL fragment modelling an equivalence relation for three classes from three different ontologies},label=fss:mapping_model_class_equivalence,captionpos=b]
EquivalentClasses( a:Person b:Persoon c:Umntu  )
\end{lstlisting}

\noindent The resultant equivalence and subsumption alignments can be saved as a multilingual OWL ontology, with the matched ontologies imported into it, or merged into one, or soft-imported by IRI.

Aforementioned SKOS can also be used for mappings, using {\tt skos:mappingRelation} \cite{SKOS:2009} and its sub-properties, to map identifiers in a pair-wise fashion between concepts in thesauri and terminologies of the same subject domain. However, such mappings can only be done between individuals of {\tt skos:Concept}, and thus requires an additional step of converting aligned {\tt skos:Concept} individuals into OWL classes to create a multilingual ontology. 
An example of SKOS-based mappings include the AGROVOC multilingual thesaurus, which uses \mbox{cross-lingual} mappings, where SKOS concepts are mapped to external vocabularies such as the Chinese Agricultural Thesaurus using {\tt skos:exactMatch} and {\tt skos:closeMatch} \cite{Caracciolo:EtAl:2013,Liang:EtAl:2005, Liang:Sini:2006}. 
Annane et al. \cite{Annane-EtAl:2016} used SKOS and the General Ontology for Linguistic Description (GOLD) \cite{Farrar:Langendoen:2003} to generate 228~000 mappings between English ontologies on BioPortal and its French equivalent.
Kuai et al.~\cite{Kuai:EtAl:2016} took two monolingual ontologies in different natural languages, automatically translated the labels of one of the ontologies to the natural language of the other ontology, and then identified mapping relationships between concept pairs. This was further examined with an algorithm to measure semantic similarity and subsequently mapped to SKOS according to its semantic similarity, using {\tt skos:exactMatch}, {\tt skos:closeMatch}, or {\tt skos:relatedMatch}.

For pair-wise mappings, if the property used for the mapping is symmetric, then a mapping from a resource in ontology O\textsubscript{1} to a resource in ontology O\textsubscript{2} consists of one axiom.
If \textit{n} is the number of classes to be mapped, then the number of correspondences required is \((n(n-1))/2 \), which quickly becomes hard to manage for multiple languages and large vocabularies or ontologies.

\subsubsection{Linguistic Mappings.}
\label{sec:modelling_multilinguality_mappingLinguistic}

\begin{figure}[t]
\centering
\includegraphics[width=0.48\textwidth]{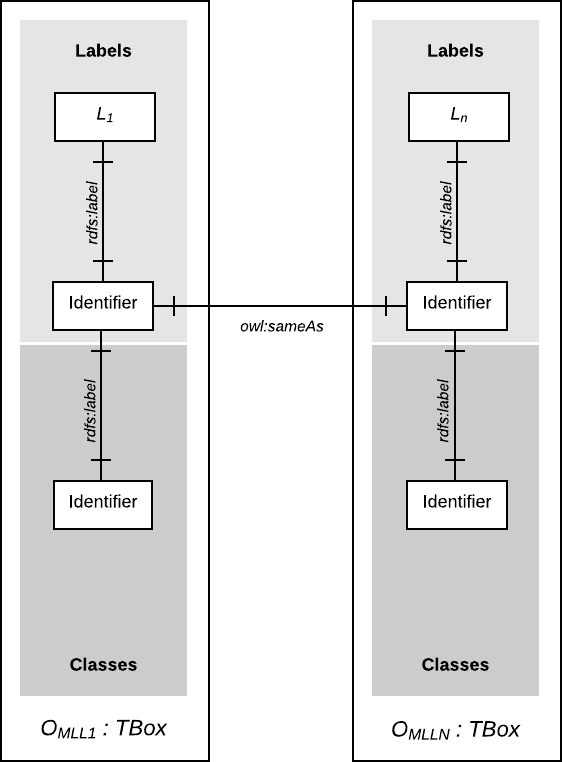}
\caption{Use of OWL as a `mapping model' in the TBox, but where equivalences are declared on the annotation values to create a multilingual ontology.}
\label{fig:model_mappingAnnotation}
\end{figure}

Where two monolingual ontologies are structurally equivalent and have labels, then pair-wise mappings could be asserted with {\tt owl:sameAs} (see Figure~\ref{fig:model_mappingAnnotation}), even though a good reason to do so may not be evident. A more elaborate scenario motivating mappings at the linguistic layer is a desire to re-use the ample existing lexical resources that have such mappings. We consider the InterLingual Index (ILI) from Global WordNet and those lexical concepts from OntoLex-Lemon to illustrate this strategy.

\begin{figure*}[t]
\centering
\includegraphics[width=0.7\textwidth]{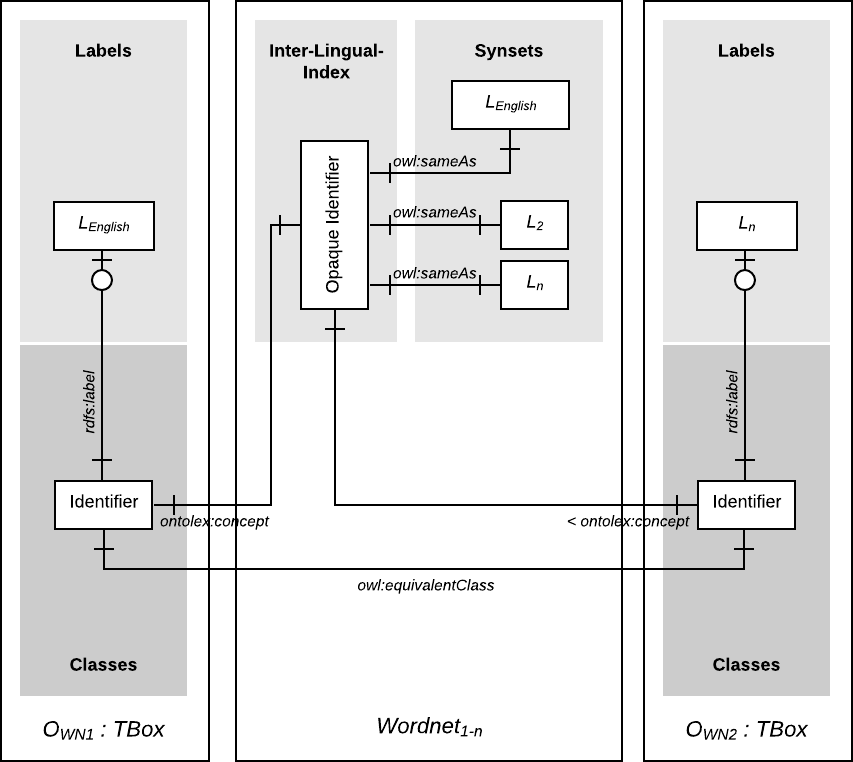}
\caption{OntoLex-Lemon is used to demonstrate the theoretical option of linguistic mappings, using Global Wordnet as the interlingua.}
\label{fig:model_mappingWordnet}
\end{figure*}

WordNet is a semantic network of synsets, where a synset is a collection of synonymous words that have an equivalent denotation; each word may be used interchangeably, depending on the context \cite{Miller:Fellbaum:2007}. 
Princeton WordNet (PWN), in English, was the first WordNet developed and eight in European languages were developed afterward \cite{Miller:Fellbaum:2007,Miller:EtAl:1990,Vossen:1997}, with other language WordNets following later. 
Each new WordNet was interlinked via PWN, with PWN used as a pivot. 
If a word was not lexicalised in PWN, however, then the concept could not be expressed. The ILI was then introduced, which is a flat list of concepts, each with a unique identifier and each synset in a WordNet is then associated with a concept from the ILI using {\tt owl:sameAs} \cite{Miller:Fellbaum:2007, GlobalWordNetBackground:2004}. This can be linked to a linguistic model, and from there to an ontology. That is, entries in ILI function as an interlingua, where each class in two monolingual ontologies can be associated with a concept from ILI using, e.g., {\tt ontolex:concept} from OntoLex-Lemon. 
The classes from each ontology that share the same ILI concept may then be aligned with {\tt owl:equivalentClass}. This approach is shown in Figure~\ref{fig:model_mappingWordnet}.
\begin{figure*}[h]
\centering
\includegraphics[width=0.83\textwidth]{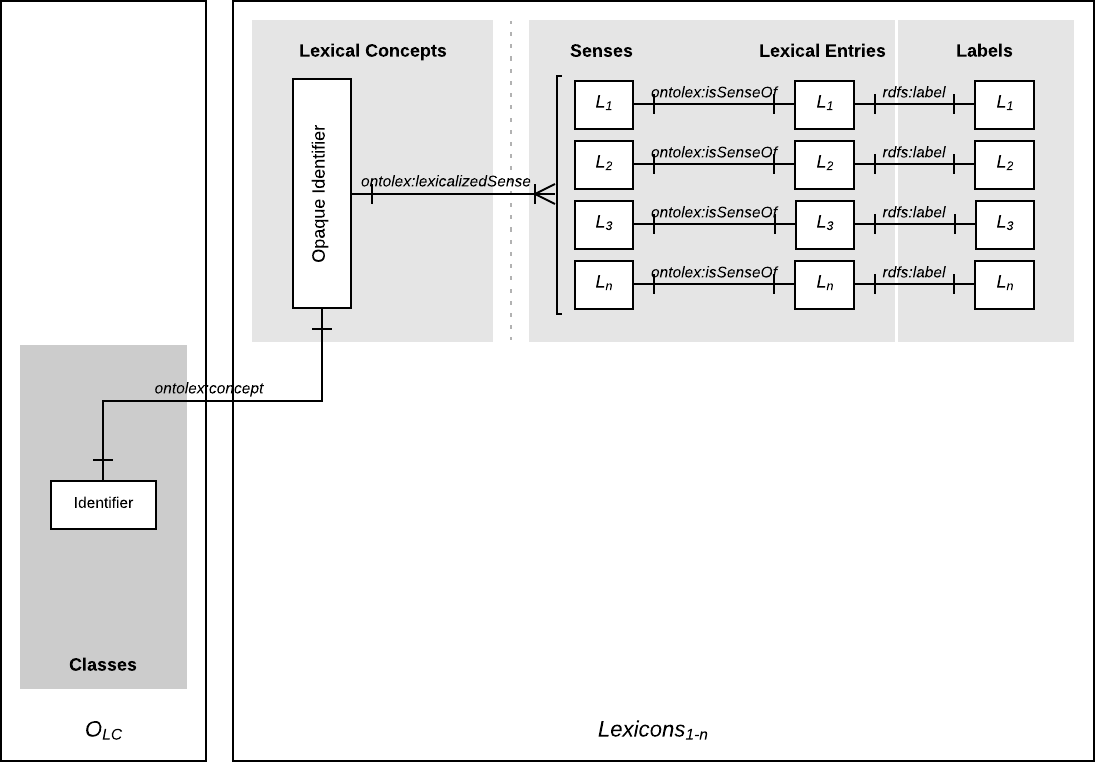}
\caption{Lexical concepts can be used as an interlingua, shown here using OntoLex-Lemon.}
\label{fig:model_mappingConcepts}
\end{figure*}
\begin{table*}[t]
\fontsize{9}{12}\selectfont
\caption{\label{table:overview_mapping_models} Overview of the Limitations of Mapping Models for Modelling Multilingualism}
\begin{tabular}{ | p{0.3\linewidth} | p{0.65\linewidth} |}
\hline
\textbf{Feature/Approach} & \textbf{Shortcoming/Limitation} \\
\hline \hline
Alignment of two or more ontologies & It requires effort to update the alignments to keep in sync with any source or target ontology changes, and to ensure the multilingual ontology stays consistent. \\ 
\hline
Different conceptualisations & When ontologies are aligned, it results in non-1-1 mappings. Custom classes may need to be created in the merged ontology to address this. \\ 
\hline
Custom alignment axioms & If custom properties are created to align ontologies, human inspection is required to identify these properties when the merged ontology is reused. \\
\hline
SKOS as a mapping model & SKOS can only be used to align instances. It cannot be used to align classes. To create a multilingual ontology from the alignments, each individual would then have to be created as a class in post-processing. \\
\hline 
Use of a `mapping hub' such as Global WordNet as the interlingua & This can only be used for those languages which have a common interlingua. In the case of Global WordNet, this would be an accompanying WordNet for that natural language. \\ 
\hline
\end{tabular}
\end{table*}
A lexical concept is associated with an OWL class using the OntoLex-Lemon {\tt ontolex:concept} property \cite{Cimiano:Mccrae:Buitelaar:2016}. In this context, a lexical concept serves as a language-independent `container' that has one or more senses from different language lexicons lexicalised to it, as shown in Figure~\ref{fig:model_mappingConcepts}. 
An example OWL fragment is given in Listing \ref{fss:mapping_model_lexical_concepts_owl}, with the corresponding lexical concept given in Listing \ref{fss:mapping_model_lexical_concepts_rdf}.

\begin{lstlisting}[language={fss},caption={OWL fragment showing an association of a class to a lexical concept},label=fss:mapping_model_lexical_concepts_owl,captionpos=b]
SubClassOf( :493Dk owl:Thing )
AnnotationAssertion(rdfs:label :493Dk "Person"@en)
AnnotationAssertion(ontolex:concept :493Dk
    ex:lexicalConcepts/000000001)
\end{lstlisting}

\begin{lstlisting}[language={fss},caption={RDF Turtle fragment of a lexical concept},label=fss:mapping_model_lexical_concepts_rdf,captionpos=b]
:lexicalConcepts/000000001
    a ontolex:LexicalConcept ;
    ontolex:lexicalizedSense
        :en/lexicalEntry_Person_sense1 ;
    ontolex:lexicalizedSense
        :nl/lexicalEntry_Persoon_sense1 ;
    ontolex:isEvokedBy :en/lexicalEntry_Person ;
    ontolex:isEvokedBy :nl/lexicalEntry_Persoon .
\end{lstlisting}

The issue with using lexical concepts or Global WordNet as an interlingua is that both require a conceptualisation to be the same for all languages. 
However, as shown by, e.g., Hirst \cite{Hirst:2014}, this is rarely universal, in particular for subject domains that rely less on concrete reality or understanding thereof, such as, say, the concept of money and the plethora of feelings.
Further, the morphological representation may be such that these forms cannot be represented adequately in lexical entries, senses, or synsets \cite{Chavula:Keet:2014,Miller:Fellbaum:2007}, as mentioned before in Section \ref{sec:modelling_multilinguality_linguistic}) and for which this option has no solution either.

As an alternative way to model linguistic cross-lingual mappings, the Translation module in \textit{lemon} and OntoLex-Lemon can be used \cite{Cimiano:Mccrae:Buitelaar:2016, Montiel-Ponsoda:EtAl:2011, Gracia:EtAl:2014}.
With this module, two lexical senses of different natural languages can reference equivalent classes in a source and target ontology.
The senses can be set as direct equivalent translations using the Translation class.
A category registry is used to model the type of translation, and in the examples given in the literature, the Translation Categories RDF Reference Schema (TRCAT) is used \cite{TranslationCategories:2021}.
Alternatively, the \textit{translation} property in OntoLex-Lemon can be used as a translation relation between  senses \cite{Cimiano:Mccrae:Buitelaar:2016}.

The benefit of mapping models is that it allows both homogenous and heterogenous resources (ontologies,  vocabularies, etc.) to be combined to create a multilingual ontology. There are limitations to this approach, however. For multilingual alignments, it cannot be assumed that the concept spaces of each language will have a 1-1 mapping.
An option here is to be either take one language as the primary language or to edit the conceptualisations of each to render a more language-independent view.
This will most likely require manual alignment to resolve.
For cross-ontology alignments, there will be local consistency but it is possible that once aligned, there will not be global consistency, and manual alignment will be required to resolve the problematic axioms.
Lastly, ontologies may evolve, and if one or more of the base ontologies are updated regularly, it will require effort to keep the multilingual ontology in sync and stay consistent.

The mapping model approach differs from the previous two approaches, in that once a merged multilingual ontology is created, the ontology then uses multilingual labels or a linguistic model for its annotations.
For example, if the instance identifiers are mapped using SKOS, then the multilingual labels approach applies to the resultant annotations.
For the proposed methods for linguistic mappings, namely Global Wordnet as the interlingua, and the use of lexical concepts, then the multilingual labels approach applies to the annotations for the former, however, the annotations use the linguistic model approach for the latter. 
For an overview of the limitations of mapping models, see Table \ref{table:overview_mapping_models}.

\section{Multilingualism in Ontologies in the BioPortal and LOV Repositories}
\label{sec:review_ontologies_bioportal}

Given the different approaches to develop a multilingual ontology, we now proceed to examine which ones are used in practice, how often, and the extent of multilingualism when it is used in an ontology. To do so, two issues need to be resolved first: the approaches can be realised in different serialisations that have to be covered somehow for a set of ontologies, and what exactly to measure of them. We proceed to the quantitative assessment afterward.

\subsection{Preliminaries}

We first identified the aspects of the serialisation on which to base search over the ontology files. To include ontology of varying age, several standards may have been used for those approaches, which complicates the details of the search. 
Annotations in RDF 1.0 use {\tt rdfs:label} and {\tt rdfs:comment}, where the former is a human-readable version of a uniform resource identifier (URI) fragment and the latter is a description of the entity represented by the URI \cite{RDFSchemaOld:1999}. 
The attribute {\tt xml:lang} is used to specify the natural language of the annotation value (as well as any other literals) \cite{RDFSchemaOld:1999}. 
To address the issues with both {\tt xml:lang} \cite{RDFIssuesTracking:2000} and the language-sensitivity required for a literal \cite{RDFModelTheoryOld:2001}, {\tt rdf:PlainLiteral}, a string combined with an optional language indicator, was introduced in RDF in 2002 \cite{RDFConceptsOld:2002}. 
Later, in 2011, this was updated again, in that the language indicator has to adhere to IETF's BCP47 \cite{Phillips:Davis:2006,Phillips:Davis:2009}, a document specifying tags to identify languages \cite{RDF1.1ConceptsOld:2011}, and in 2012, the {\tt rdf:langString}  datatype was specified for language-tagged literals \cite{RDF1.1ConceptsOld:2012}. In parallel, for OWL, the annotation value of {\tt rdfs:label} could only be data literals in the original OWL \cite{OwlSemantics:2004}, but for OWL 2 it can be a data literal, URI, or individual \cite{OwlStructural:2012}.

Based on these variations, we specified the search criteria for the multilingual labels approach as any class or property in OWL that uses an {\tt rdfs:label} with a language-tagged string or an {\tt xml:lang} attribute.
To identify those multilingual ontologies using a URI for the annotation value, the search criteria was determined as any {\tt rdfs:label} which had a URI as the annotation value with a local namespace.
For the linguistic model approach, the search criteria consisted of the namespaces of the following models: {\tt ontolex}, {\tt linginfo}, {\tt lexonto}, {\tt lexinfo}, {\tt gold} and {\tt lemon}, with manual analysis of each ontology thereafter for its multilingual aspects.
Similarly for the mapping model approach, the search criteria was specified as {\tt owl:sameAs} and the {\tt skos} namespace, with manual analysis thereafter.

For metrics, we devised a notion of {\em coverage} and of {\em completeness} specifically for the TBox of the ontology that we are interested in\footnote{Also, the ABox introduces other issues in multilingualism. For instance, it can contain individuals with monolingual proper names, such as a book title, but  this does not necessarily extend to all languages. For example, for people's names, `Bill Gates' renders as \textit{uBill Gates} in isiXhosa, and accurately determining the named individuals in an ontology requires manual inspection.}. 
Coverage is defined as follows. 

\begin{definition}[\textbf{Coverage (Cov)}]
\label{definition_tc}
V\textsubscript{C}, V\textsubscript{OP} and V\textsubscript{DP} are the set of classes, object properties, and data properties, respectively, as defined in the OWL 2 Specification \cite{OwlStructural:2012}.
{\em Coverage (Cov)} is the sum of the number of elements in V\textsubscript{C}, V\textsubscript{OP} and V\textsubscript{DP}, prior to any ontology imports.
If another namespace is used for a class or property without being imported using \verb+owl:imports+, then these elements are included in {\em Cov}.
\end{definition}

We exclude imported ontologies as their multilinguality may differ from the ontology doing the importing, with the result that coverage may be skewed, particularly when only a few elements may be used from each imported ontology. 

Language-specific completeness is defined as follows.

\begin{definition}[\textbf{Language-Specific Completeness (LCom\textsubscript{lang})}]
\label{definition_lcov}
LCom\textsubscript{lang} for a natural language {\em lang} is the sum of language-specific literals for each element in V\textsubscript{C}, V\textsubscript{OP}, and V\textsubscript{DP} divided by Cov. 
\end{definition}

\noindent These metrics are illustrated in Example 1.

\begin{example}
\label{example_lcov}
Let us assume there is some ontology \textit{O} that has 50 classes, 10 object properties, and 5 data properties. It uses {\tt rdfs:label} for language-tagged labels for each of the classes and properties, for three natural languages, being English, French, and German. Labels are provided for all classes and object properties for English; only for the classes with `fr' tagged literals for French; and only 30 of the classes have labels with `de' tagged literals for German.

The first metric to compute is coverage: 
\begin{equation*}
\mathrm{Cov} {}  = |V_{C}| + |V_{\mathrm{OP}}| + |V_{\mathrm{DP}}| 
                 = 50 + 10 + 5 
                 = 65 
\end{equation*}                
The language-specific completeness for each of the three languages is then computed as follows:
\begin{equation*}
\begin{aligned}                
\mathrm{LCom}_{\mathrm{en}} {} & = (|{V_{C}}|_{labels} + {|V_{\mathrm{OP}}}|_{labels} + {|V_{\mathrm{DP}}}|_{labels})\;/\;\mathrm{Cov} * 100 \\
                & = (50 + 10 + 0)\;/\;65 * 100 = 92,3\% \\ 
\mathrm{LCom}_{\mathrm{fr}} {} & = (50 + 0 + 0)\;/\;65 * 100  = 76,9\% \\ 
\mathrm{LCom}_{\mathrm{de}} {} & = (30 + 0 + 0)\;/\;65 * 100  = 45,2\% 
\end{aligned}
\end{equation*}
\end{example}

\noindent The computed LCom\textsubscript{lang} values lead to the notion of a primary language of an ontology:

\begin{definition}[\textbf{Primary Language (PL)}]
\label{definition_pl}
The PL of the ontology is the natural language with the highest percentage LCom\textsubscript{lang}.
\end{definition}

At least in theory it is possible that more than one natural language is the primary language in an ontology if the percentages are the same or at least very similar.  
If more than one primary language is supported for an ontology, it suggests that it is a language-independent ontology. To  ascertain this, an ontology would need to be manually inspected, focussing on the identifier type and the labels used, particularly the consistency thereof for each of its languages.

\subsection{Preparation of the Datasets from the repositories}

Since there are multiple ways to model multilinguality and various ontology formats, we first describe in detail the concrete processes for dataset creation. We use two well-known ontology repositories, BioPortal and the LOV dataset for this.

\paragraph{Preparation of the BioPortal Dataset}
\label{sec:review_ontologies_requirements}

The following steps were undertaken:
\begin{compactenum}
\item Using BioPortal's REST API, both a list of ontologies and categories were downloaded using the \textit{/ontologies} and \textit{/categories} endpoints on the 28/04/2022.
The data were saved as JSON files, which were then imported into a MySQL database.
The result was a listing of 981 ontologies.
Using a script, the metadata of 30 ontologies were retrieved at a time, using the \textit{/ontologies/\{acronym\}/categories} and \textit{/ontologies/\{acronym\}/latest\_submission} endpoints.
The list of categories each ontology was linked to, the ontology language used, the status and the description were then saved to the database.
\item The dataset was then filtered on OWL ontologies only (other encoding options were UMLS, SKOS, and OBO). This resulted in 730 ontologies.
\item The dataset was filtered on those OWL ontologies with a status of `production' (other statuses are `alpha', `beta', `retired' and \textit{null}), which is assumed to be the most mature version of the ontology. 268 ontologies remained after this selection. 
\end{compactenum}
Of the remaining ontologies, two ontologies returned an empty result and were excluded from evaluation.  
This resulted in a final dataset of 266 ontologies from BioPortal. 

The methodology for review comprised the following: each ontology file was loaded in memory, with a script used to loop through each line therein.
Any line which mapped to one or more of the search criteria was then saved to the database. 
These saved rows were then manually inspected to confirm their multilingual aspects.

\paragraph{Preparation of the LOV Dataset}
\label{sec:review_ontologies_requirements_lov}

The following steps were undertaken:
\begin{compactenum}
\item LOV provides an API, with an endpoint to retrieve the list of vocabularies: \textit{/api/v2/vocabulary/list}. On 29/04/2022, the list of vocabularies were retrieved from this URL, and the contents saved as a JSON file.
\item The JSON file was then imported into the same MySQL database. The result was a listing of 773 vocabularies.
\item Using a script, the contents of each vocabulary were retrieved using a cURL command, with the Content-Type of the header set to {\tt application/rdf+xml}. RDF/XML was selected as it is the mandatory serialisation format of OWL ontologies.
\item For those vocabularies that returned an HTTP status code of 4*, the vocabularies were manually reviewed on LOV. Using the VAPOUR hyperlink provided on LOV, if the linked data validator VAPOUR\footnote{\url{http://uriburner.com:8000/vapour}} returned no results, then no further action was taken.
\item For those vocabularies that returned an HTTP status code of 0, the vocabularies were manually reviewed using their URI. With the exception of one vocabulary, all vocabularies were unreachable due to timeout or DNS issues. No further action was taken for these vocabularies.
\item For those vocabularies that returned an HTTP status code of 200, each RDF document was then manually opened. If the file presented a web document in HTML, then the vocabulary was manually reviewed on LOV, following the same steps in Point 4. If VAPOUR provided a file for a different Content-Type (options included {\tt xhtml+xml} and {\tt text/turtle}) and an RDF document was available, then the web document was updated to this file. If no file could be returned by VAPOUR, then the N3 serialisation was downloaded from LOV.
\end{compactenum}

\noindent Of the 773 vocabularies, the following was the status at time of evaluation:

\begin{compactitem}[-]
\item HTTP status code of 0: 75 vocabularies, including one that was manually reviewed and found to have a valid RDF document.
\item HTTP status code of 3*: 23 vocabularies
\item HTTP status code of 4*: 125 vocabularies
\item HTTP status code of 5*: 27 vocabularies
\item HTTP status code of 200: 523 vocabularies
\end{compactitem}

\noindent Regarding the vocabularies with a status code of 200, two  returned an RDF document showing only namespaces, of which one was Schema.org, and 103 did not have an RDF document served by their URI using content negotiation. The latter included the Data Catalog Vocabulary (DCAT), three vocabularies from Dublin Core, the ISO 25964 SKOS extension, JSON Schema in RDF (a vocabulary provided by a W3C working group), Lexvo.org Ontology, W3C Shapes Constraint Language (SHACL) Vocabulary, and the RDFa Vocabulary for Term and Prefix Assignment.

This last step left a dataset of 521 vocabularies for assessment. 
The same evaluation process as that for BioPortal was applied to these vocabularies.

\begin{table*}[t]
\begin{threeparttable}
\fontsize{8}{12}\selectfont
\caption{\label{table:multilingual_ontologies} Classification of Identified Multilingual Ontologies in BioPortal}
\begin{tabular}{@{}p{7.5cm}||M{2.5cm}|M{2.5cm}||M{3.5cm}@{}}
\hline
\textbf{Ontology} & \textbf{Primary Language(s)} & \textbf{Other Language(s)} & \textbf{Modelling Option} \\
\hline
\hline
Animal Trait Ontology for Livestock (ATOL) & en, fr & - & Multilingual Labels \\
\hline
CIDOC Conceptual Reference Model (CIDOC-CRM) & de, el, en, fr, pt, ru, zh & - & Multilingual Labels \\
\hline
Cell Ontology (CL) & en & zh & Multilingual Labels \\
\hline
WHO COVID-19 Rapid Version CRF Semantic Data Model (COVIDCRFRAPID) & en & pt-br & Multilingual Labels \\
\hline
Data Catalog Vocabulary (DCAT-FDC) & en & ar, da, el, cs, es, fr, it, ja & Multilingual Labels \\
\hline
VEuPathDB Ontology (EUPATH) & en & fr, pt & Multilingual Labels \\
\hline
Clinical LABoratory Ontology (LABO) & en & fr & Multilingual Labels \\
\hline
The MOSAiC Ontology (MOSAIC) & en & es & Multilingual Labels \\
\hline
Nanbyo Disease Ontology (NANDO) & en, ja & - & Multilingual Labels \\
\hline
Ontology for Biomedical Investigations (OBI) & en & zh & Multilingual Labels \\
\hline
Ontology for Biobanking (OBIB) & en & zh & Multilingual Labels \\
\hline
Ontology of Chinese Medicine for Rheumatism (OCMR) & en & zh & Multilingual Labels \\
\hline
Ontology of Units of Measure (OM) & en & ja & Multilingual Labels \\
\hline
Emergency Care Ontology (ONTOLURGENCES) & fr & en & Multilingual Labels \\
\hline
The Prescription of Drugs Ontology (PDRO) & en & fr & Multilingual Labels \\
\hline
Radiology Lexicon (RADLEX) & de, en & - & Multilingual Labels \\
\hline
Sequence Ontology (SEQ) & en & it & Multilingual Labels \\
\hline
Viral Disease Ontology Trunk (VDOT) & en & de & Multilingual Labels \\
\hline
\hline
\end{tabular}
\end{threeparttable}
\end{table*}

\subsection{Identification of Multilingual Ontologies}
\label{sec:review_ontologies_identification}

The results of the dataset from BioPortal is included in Table~\ref{table:multilingual_ontologies}, with both the primary language(s) and any other natural languages indicated. Of the 266 ontologies, 18 ontologies were identified as multilingual. Their Cov and LCom\textsubscript{lang} metrics are included in Table~\ref{table:multilingual_ontologies_metrics}. For example, for OM, LCom\textsubscript{en} and LCom\textsubscript{ja} is 33.9\% and 2.0\% respectively, where `ja' is the language code for Japanese. 
If we exclude from consideration all natural languages with 5\% or less completeness in an ontology, then CL, EUPATH, MOSAIC, OBI, OBIB, OCMR, and OM no longer qualify as a multilingual ontology.
The extent of multilingualism of the ontologies, including all natural languages regardless of their completeness, amounts to 6.77\% of BioPortal's production ontologies in the dataset. Excluding those natural languages with $\leq 5\%$ completeness, then multilingualism is only 4.14\%. 

\begin{table*}[t]
\begin{threeparttable}
\fontsize{6}{10}\selectfont
\caption{\label{table:multilingual_ontologies_metrics} Label Metrics: Language-specific Coverage (LCom\textsubscript{lang}) Compared to Total Element Count (Cov) for that Ontology, for each BioPortal ontology identified as multilingual.}
\begin{tabular}{@{}p{1.55cm}||M{0.9cm}|M{0.6cm}|M{0.6cm}|M{0.6cm}|M{0.6cm}|M{0.6cm}|M{0.6cm}|M{0.6cm}|M{0.6cm}|M{0.6cm}|M{0.6cm}|M{0.6cm}|M{0.6cm}|M{0.6cm}|M{0.6cm}@{}}
\hline
\textbf{Ontology} & \textbf{Cov} & \textbf{ar} & \textbf{cs} & \textbf{da} & \textbf{de} & \textbf{el} & \textbf{en} & \textbf{es} & \textbf{fr} & \textbf{it} & \textbf{ja} & \textbf{pt} & \textbf{pt-br} & \textbf{ru} & \textbf{zh} \\
\hline
\hline
ATOL & 2 352 & - & - & - & - & - & 100\% & - & 100\% & - & - & - & - & - & - \\
\hline
CIDOC-CRM & 372 & - & - & - & 92.5\% & 87.4\% & 99.7\% & - & 87.4\% & - & - & 87.4\% & 87.1\% & 90.9\% & - \\
\hline
CL & 16 846 & - & - & - & - & - & 1.3\% & - & -& -  & - & - & - & - & 0.1\% \\
\hline
COVIDCRFR. & 407 & - & - & - & - & - & 78.9\% & - & - & - & - & - & 53.8\% & - & - \\
\hline
DCAT-FDC & 39 & 41.0\% & 89.7\% & 92.3\% & - & 41.0\% & 94.9\% & 89.7\% & 41.0\% & 94.9\% & 41.0\% & - & - & - & - \\
\hline
EUPATH & 4 184 & - & - & - & - & - & 15.5\% & - & 0.02\% & - & - & 0.1\% & - & - & - \\
\hline
LABO & 204 & - & - & - & - & - & 90.7\% & - & 10.3\% & - & - & - & - & - & - \\
\hline
MOSAIC & 282 & - & - & - & - & - & 37.9\% & 3.2\% & - & - & - & - & - & - & - \\
\hline
NANDO & 2 733 & - & - & - & - & - & 100\% & - & - & - & 100.\% & - & - & - & - \\
\hline
OBI & 4 733 & - & - & - & - & - & 25.7\% & - & - & - & - & - & - & - & 0.1\% \\
\hline
OBIB & 1 949 & - & - & - & - & - & 32.1\% & - & - & - & - & - & - & - & 0.2\% \\
\hline
OCMR & 3 471 & - & - & - & - & - & 5.0\% & - & - & - & - & - & - & - & 1.3\% \\
\hline
OM & 833 & - & - & - & - & - & 33.9\% & - & - & - & 2.0\% & - & - & - & - \\
\hline
ONTOLURG. & 10 092 & - & - & - & - & - & 28.2\% & - & 99.0\% & - & - & - & - & - & - \\
\hline
PDRO & 239 & - & - & - & - & - & 74.1\% & - & 62.3\% & - & - & - & - & - & - \\
\hline
RADLEX & 46 813 & - & - & - & 46.3\% & - & 46.5\% & - & - & - & - & - & - & - & - \\
\hline
SEQ & 5 & - & - & - & - & - & 80.0\% & - & - & 80.0\% & - & - & - & - & - \\
\hline
VDOT & 208 & - & - & - & 37.0\% & - & 29.8\% & - & - & - & - & - & - & - & - \\
\hline
\hline
\end{tabular}
\end{threeparttable}
\end{table*}

\begin{table}
\begin{threeparttable}
\fontsize{8}{12}\selectfont
\caption{\label{table:multilingual_datasets} Comparison of BioPortal to LOV}
\begin{tabular}{@{}p{5cm}||M{1.3cm}|M{1.3cm}@{}}
\hline
\textbf{} & \textbf{BioPortal} & \textbf{LOV} \\
\hline
\hline
Number of ontologies in dataset & 266 & 521 \\
\hline
\hline
\textbf{Completeness $>0\%$} &  &  \\
\hline
Multilingual ontologies  & 18 & 82 \\
\hline
Percentage of multilingualism & 6.77\% & 15.74\% \\
\hline
Total Cov of all multilingual ontologies & 95 762
& 14 644 \\
\hline
Mean of Total Cov & 5 320.11 & 178.59 \\
\hline
Median of Cov & 1 391 & 66 \\
\hline
\hline
\textbf{Completeness $>5\%$} & 
&  \\
\hline
Multilingual ontologies  & 11 & 74 \\
\hline
Percentage of multilingualism & 4.14\% & 14.20\% \\
\hline
Total Cov of all multilingual ontologies & 63 464 & 9 203 \\
\hline
Mean of Total Cov & 5 769.46 & 124.37 \\
\hline
Median of Cov & 372 & 63 \\
\hline
\hline
\end{tabular}
\end{threeparttable}
\end{table}

\begin{figure}[h]
\centering
\includegraphics[width=0.40\textwidth]{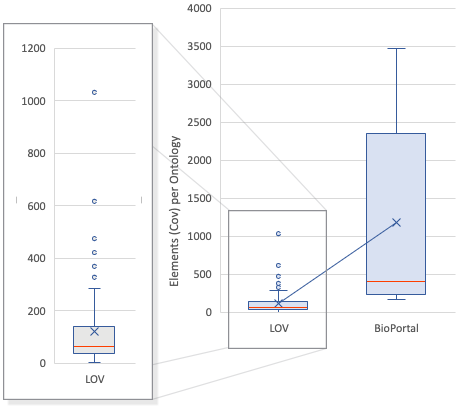}
\caption{Sizes of multilingual ontologies in LOV compared BioPortal (excluding the outliers ONTOLURGENCES and OM from the BioPortal dataset, and OBI from the LOV dataset, due to visualisation constraints on the graph). A zoomed-in view of LOV is provided on the left. The connecting line between LOV and BioPortal shows the mean between both, and the red line in each is the median.}
\label{fig:graph_boxplot}
\end{figure}

For the LOV dataset, of the 521 vocabularies, 82 multilingual vocabularies were identified, amounting to 15.74\%  multilingual. The Cov and LCom\textsubscript{lang} metrics were computed on them as well, but due its size, we made them available as online supplementary material at \url{https://fynbosch.com/article-2022-bioportal-lov-review}. If we exclude those natural languages with $\leq 5\%$ completeness, then multilingualism reduces to 14.20\%.

The top 5 languages in LOV is English first, followed by Italian, French, Spanish and German.
The smallest vocabulary is TI, with a Cov of 4, and the biggest is OBI\footnote{In LOV, OBI has the acronym of `OBO'; to avoid confusion, we use the acronym OBI as it is the ontology's known acronym.}, with a Cov of 4731.
However, OBI is also one of the vocabularies that does not qualify as multilingual if we exclude natural languages with $\leq 5\%$ completeness. 
If vocabularies with a completeness greater than 5\% are considered, then the biggest vocabulary is KM4C with a Cov of 1033.
The vocabularies that support the most number of natural languages is BTO, LINGVO, and MIL, with 16.
Again, if we exclude for $\leq 5\%$ completeness, then only BTO supports 16 languages, with LINGVO and MIL dropping to 6 and 3 respectively.
The number of vocabularies for each natural language (with $>5\%$ completeness) is in alphabetical order: Afrikaans~(5), Arabic~(2), Catalan~(1), Chinese~(1), Czech~(5), Danish~(3), Dutch~(9), English~(74), Estonian~(1), Farsi~(1), Finnish~(2), French~(27), German~(19), Greek~(1), Italian~(31), Japanese~(2), Korean~(3), Portuguese~(8), Romanian~(5), Russian~(6), Slovak~(1), Spanish~(20), Swedish~(6), and Turkish~(1).

\begin{table}
\begin{threeparttable}
\fontsize{8}{12}\selectfont
\caption{\label{table:language_distribution} Number of vocabularies (Vocabs) with 2--16 natural languages in LOV, with language completeness of $>0\%$ compared to $>5\%$}
\begin{tabular}{@{}p{3cm}||M{2.3cm}|M{2.3cm}@{}}
\hline
\textbf{Natural Languages} & $|$\textbf{Vocabs}$|$ with $>0\%$  & $|$\textbf{Vocabs}$|$ with $>5\%$ \\
\hline
\hline
2 & 65 & 58 \\
\hline
3 & 3 & 3 \\
\hline
4 & 2 & 2 \\
\hline
5 & 2 & 2 \\
\hline
6 & 1 & 2 \\
\hline
7 & 0 & 0 \\
\hline
8 & 1 & 1 \\
\hline
9 & 0 & 0 \\
\hline
10 & 0 & 0 \\
\hline
11 & 5 & 5 \\
\hline
12 & 0 & 0 \\
\hline
13 & 0 & 0 \\
\hline
14 & 0 & 0 \\
\hline
15 & 0 & 0 \\
\hline
16 & 3 & 1 \\
\hline
\hline
\end{tabular}
\end{threeparttable}
\end{table}

By number and percentage of ontologies, LOV seems much more multilingual than BioPortal. It is not that simple, which can be seen from the data in Table~\ref{table:multilingual_datasets} and the boxplot in Figure~\ref{fig:graph_boxplot}. We calculated both the mean and median of the multilingual ontologies in each dataset, summing the Cov metric of each multilingual ontology, to determine if the ontologies and vocabularies between BioPortal and LOV are similarly sized. The mean of the BioPortal multilingual dataset is 5~320.11 (95~762/18), and the mean of the LOV multilingual dataset is 178.59 (14~644/82). That is, while the percentage of multilingualism is higher in the LOV dataset, those ontologies are much smaller. Whether they are more multilingual because they are smaller, or for another reason, cannot be determined from the data. 

The language distribution for LOV is shown in Table~\ref{table:language_distribution}.

\section{Management Options for Multilingual Ontologies}
\label{sec:managing_multilinguality}

Multilingual ontology management is the management of the multilingual aspects of an ontology in an Ontology Development Environment (ODE), whichever form the tooling may take. To be able to do this effectively, such tools need to meet a set of requirements at least to some extent. 
Of the requirements that were identified, three broad categories emerged: \textit{multilinguality}, \textit{interaction} and \textit{system architecture}.
For \textit{multilinguality}, this concerns the ability to implement and manage the different modelling options beyond that of multilingual {\tt rdfs:label}s.
\textit{Interaction}, refers to the need for an end-user to at least view an ontology localised to a specific natural language.
In addition to this, OWL verbalisation was also identified as a requirement, in a similar vein to that provided by NORMA, a modelling tool for Object-Role Modelling (ORM)\footnote{ORM is a fact-based modelling language formalised with a logic-based reconstruction in first order predicate logic. Using controlled natural languages, verbalisations are visualised for objects and their roles (relations) \cite{Halpin:Curland:2006}.}. 
This can be similarly applied to an OWL user.
Using a controlled natural language for a language allows easier evaluation of those language-specific labels, particularly when used in combination with the different modelling options for multilinguality.

For \textit{system architecture}, versioning is a necessity, particularly if the ontology is used by other resources. 
Although this is not unique to multilingual ontologies, it is more pressing since more can change and has to be maintained in a multilingual setting. There is also an expectation that multiple authors edit the same OWL `codebase', be it the file as the ontology size starts to increase and needing more domain experts, or the related language files, since then also expert linguistic and translation roles may become a necessary.
Lastly, guiding principles for scientific data management and stewardship, known as FAIR, were proposed by \cite{Wilkinson:EtAl:2016} in 2016.
Since then, the community uptake has been rapid \cite{Jacobsen:EtAl:2020}, with the result that the FAIR principles\footnote{\url{https://www.go-fair.org/fair-principles/}} have also been included with the set of requirements identified for multilingual ontology management.

In order to compare tooling support for developing and maintaining multilingual ontologies, we will use the following specific set of requirements:
\begin{compactenum}
\item \textit{Linguistic Model.}
One or more linguistic models should be associable with an editor, ideally with a user interface provided to manage each respective model.
\item \textit{Lexicons.} Like that of linguistic models, lexicons should also be able to be associated with an editor, with a user interface provided to view and/or manage them.
\item \textit{Mappings.} The bridging axioms/mappings between one or more monolingual ontologies (localisations) should be manageable within a multilingual ontology.
Although semantic mappings are typically set in a separate document, a user interface specifically to manage them is deemed useful.
\item \textit{Localisation and viewing thereof.} An ontology can be localised on both its lexical and semantic layers. If lexical only, the underlying ontology does not change.
It should be possible to view the localisation of a multilingual ontology.
\item \textit{Verbalisation.} The verbalisations of entities and axioms using a controlled natural language can bridge the gap between knowledge engineers and domain/language experts \cite{Jarrar:Keet:Dongilli:2006}, therefore verbalisations of OWL statements using a controlled natural language should be possible for each of the languages supported in the multilingual ontology.
\item \textit{Synchronisation.} This refers to synchronising localised ontologies to any changes in the external ontology.
\item \textit{Multi-Role Collaboration.}
In the multilingual ontology development cycle, in addition to the knowledge engineer and domain expert roles, the language expert role is also required \cite{Keet:2018, Dragoni-EtAl:2013}.
\item \textit{Versioning.}
Versioning of the ontology should be supported in the ODE beyond that of versioned IRIs.
Versioning can be state-based or change-based \cite{Dragoni:Ghidini:2012}.
For the latter, this should include a history of changes to an ontology, with provenance for each action.
For the former, snapshots of the dataset should be able to be taken, with users able to view a specific snapshot, either at resource-level or maintain that snapshot-state as they navigate through the ontology.
\item \textit{FAIR Principles.} The ODE should make it possible for the FAIR principles to be supported, where applicable, for an ontology.
Of the ten principles, we assume that all but F4 and A2 apply to ontologies and RDF documents; the others are excluded as they apply to the ontology, not the ODE. However, the ontology editor should provide a means to manage the metadata. For instance, for F1's unique identifiers, they should be assignable by the ontology editor (but to be globally unique, there is a dependency on A1), and also F3 and I1 should be supported by the ontology editor, as should F2, I2, I3 and R1 (though noting that making a functionality available may not make the ontology engineer use it to fulfil each principle).
\end{compactenum}

Requirements 1--3 relate to \textit{multilinguality}, 4 and 5 to \textit{interaction}, and 6--9 to \textit{system architecture}.

\subsection{Review of Tools for Multilingual Ontology Management}
\label{sec:manage_review}
A list of ODEs, plugins, and systems were identified in the literature to support (or in some instances, possibly support) the management of multilingual ontologies.
They are briefly described first. 

General ontology editors included are Prot\'eg\'e, the NeOn Toolkit, Moki, and OntoWiki. Prot\'eg\'e for OWL is a free and open-source ODE available for both desktop and web \cite{Musen:2015}, which remains under active development; the current desktop version released in 2019 is 5.5.0. The NeOn Toolkit (incl. the LabelTranslator plugin) was an open-source ODE consisting of thirty plugins \cite{Adamou:EtAl:2012, Erdmann:Walter:2012, Espinoza:Gomez-Perez:Montiel-Ponsoda:2009}, and last updated in December 2011. MoKi is a wiki-based conceptual modelling tool for ontologies and business process models \cite{Dragoni-EtAl:2013, Ghidini:EtAl:2009}, but at time of writing, the website was not available. OntoWiki is an open-source wiki-type platform for managing semantic content \cite{Martin:EtAl:2011}, which  remains under active development, with the last update in July 2017. TopBraid Composer (Maestro edition)  \cite{TopBraid:2020} was considered, but as access to the free edition is no longer available, its was excluded from the comparison.

Broadening the scope, also two Prot\'eg\'e plugins were included:  NaturalOWL and OntoLing. NaturalOWL is an open-source natural language generation system for version 4.1 of Prot\'eg\'e, and was last updated in 2014 \cite{Androutsopoulos:Lampouras:Galanis:2014}. OntoLing is an open-source plugin to support the linguistic enrichment of ontologies \cite{Pazienza:Stellato:2005}, which was last updated in 2006 and is compatible up to version 3.3.1 of Prot\'eg\'e. Further, we included VocBench 3, which is an open-source web platform for the management of RDF datasets \cite{Stellato:EtAl:2020}. It remains under active development and at time of writing, was at version 3.7.0, provides ``an almost Full coverage of OWL2 expressions", and is primarily intended for managing thesauri, vocabularies and lightweight ontologies \cite[p.17]{Stellato:EtAl:2020}.

In addition to this, the following tools were identified as possible candidates, but upon closer inspection, they are primarily for lexicography and vocabulary management, which is out of scope: 
(1) LexO, a web editor used for the creation of linked data resources using OntoLex-Lemon
\cite{Bellandi:EtAl:2017, Bellandi:Giovannetti:2020},
(2) Tedi, a multilingual terminology editor which uses a shared ontology to link different terms \cite{Papadopoulou:Roche:2018, Tedi:2020}, and
(3) TemaTres, a web application to manage controlled vocabularies \cite{TemaTres:2021}.

Of the tools examined, some were no longer available for download (notably MoKI) so assessments were done from information in the literature and available screenshots.
All other tools were locally installed for the purpose of inspection.
Each tool was then examined against afore-mentioned requirements.

\begin{table*}[t]
\begin{threeparttable}
\fontsize{9}{12}\selectfont
\caption{\label{table:multilingual_ontology_management} Feature Assessment of Selected Tools; see text for details}
\begin{tabular}{@{}p{2.1cm}||M{1.05cm}||M{1.05cm}|P{1.05cm}|M{1.05cm}||M{1.05cm}|M{1.05cm}||M{1.05cm}|M{1.05cm}|M{1.05cm}|M{1.05cm}@{}}
\hline
\textbf{ODE} & \textbf{OWL~2} & \multicolumn{3}{c||}{\textbf{Multilinguality}} & \multicolumn{2}{c||}{\textbf{Interaction}} & \multicolumn{4}{c}{\textbf{System Architecture}} \\
\hline
& & (1) & (2) & (3) & (4) & (5) & (6) & (7) & (8) & (9) \\
& & Ling. Model & Lexicons & Mapp. & Localis. & Verbal. & Synchr. & Multi-Role Collab. & Version. & FAIR\\
\hline
\hline
MoKi            & - & - & - & \checkmark & - & - & - & \checkmark & - & N/A \\
\hline
NaturalOWL      & \checkmark & - & \checkmark & - & - & \checkmark & - & - & - & 3 \\
\hline
NeOn            & \checkmark & \checkmark & \checkmark & - & - & - & - & \checkmark & \checkmark & 8 \\
\hline
OntoLing        & - & - & \checkmark & - & - & - & - & - & - & 2 \\
\hline
OntoWiki        & - & - & - & - & - & - & - & - & \checkmark & 8 \\
\hline
Prot\'eg\'e 5.x     & \checkmark & - & - & - & - & - & - & - & - & 8 \\
\hline
VocBench 3      & Some & \checkmark & \checkmark & \checkmark & - & - & - & \checkmark & \checkmark & 8 \\
\hline
\hline
\end{tabular}
\end{threeparttable}
\end{table*}

The outcomes of the tool assessment against the requirements are shown in Table~\ref{table:multilingual_ontology_management}, with the column headings aligning to that of the requirements, with an additional column to indicate whether the tool supports OWL~2. We briefly discuss selected features for applicable tools.
\begin{itemize}
\item
\textbf{Linguistic models:}  A linguistic model is integrated with both NeOn and VocBench~3, with user interfaces provided for both.
For the former, this is by way of the LabelTranslator plugin, which supports the Linguistic Information Repository (LIR) model \cite{Adamou:EtAl:2012, Espinoza:Gomez-Perez:Mena:2008}; the latter provides support for OntoLex-Lemon \cite{Stellato:EtAl:2020}.
\item
\textbf{Lexicons:} For NaturalOWL, an interface is provided to both view and manage the lexicon entries; however, only the English and Greek languages are supported \cite{Androutsopoulos:Lampouras:Galanis:2014}.
For NeOn, the lexical entries can be managed separately from the ontology entities, and, in theory, any number of natural languages can be supported \cite{Espinoza:Gomez-Perez:Mena:2008}.
For OntoLing, an interface is provided to load and view an external linguistic resource from WordNet or a bilingual dictionary from the DICT Interface for Bilingual Dictionaries \cite{Pazienza:Stellato:2005}.
In VocBench 3, lexicons and lexical entries are manageable via the user interface, with any number of natural languages able to be supported \cite{Stellato:EtAl:2020}.
\item
\textbf{Mappings:} For multilingual mappings, MoKi's dictionary-based translation (DBT) service is used \cite{Dragoni:Ghidini:Bosca:2013}, and for VocBench 3, SKOS is used as the mapping model \cite{Stellato:EtAl:2020}.
\item
\textbf{Localisations:} Lexical localisations in the form of language-tagged {\tt rdfs:label} strings are typically shown in Prot\'eg\'e.
If a particular language has been set, and the user changes their operating system localisation to that language, Prot\'eg\'e will default to this language.
It is not possible to visualise any language-specific conceptualisations in the available tools for an ontology.
\item
\textbf{Verbalisations:} Only NaturalOWL offers this feature, in both English and Greek \cite{Androutsopoulos:Lampouras:Galanis:2014}. Ontology verbalisation and, more broadly, natural language generation from ontologies, is a well-known task, but this is typically done in `third-party' tools or proof-of-concept tools \cite{BouayadAgha14,Safwat16}, such as \cite{KXK17} for isiZulu and in application scenarios like multiple-choice question generation \cite{Kurdi20}. 
\item
\textbf{Multi-Role Collaboration:} Multi-role collaboration is supported by MoKi, including the language expert role \cite{Dragoni-EtAl:2013}.
For NeOn, there is a Collaboration Support plugin which only provides support for the Viewer, Subject Expert and Validator roles \cite{NeOnCollaboration:2011}.
VocBench 3 supports multi-role collaboration with role-based access control for areas, subjects and scopes \cite{Stellato-EtAl:2017}.
Project-local roles provide support for Ontology Editors, Thesaurus and Lexicography Editors, and Validators.
For Lexicography Editors, this role can be limited to edit selected natural languages only.
Collaborative Prot\'eg\'e \cite{Tudorache:2009} is available for Prot\'eg\'e 3, but as it is not compatible with newer versions, it has not been considered here. The more recent WebProt\'eg\'e is intended for collaboration, which is mainly multiple people in general without designated roles \cite{Horridge19}.

\item
\textbf{Versioning:} OntoWiki provides support for change-based versioning of RDF data, with all activities tracked, and provenance provided for each changeset \cite{Martin:EtAl:2011, Frischmuth:EtAl:2015}.
NeOn provides support for change-based versioning using the Change Capturing plugin \cite{NeOnChange:2011}.
VocBench 3 provides support for both change-based and state-based versioning: the former is supported via a history, and for the latter, snapshots of the ontology are also supported \cite{Stellato:EtAl:2020, Stellato-EtAl:2017}.
A user can then switch their user interface view between snapshots globally or at resource-level.

\item
\textbf{FAIR Principles:} NaturalOWL supports I1, I2 and I3, and OntoLing I1 and R1.
NeOn, OntoWiki, Prot\'eg\'e and VocBench 3 provide support for all eight principles: F1, F2, F3, A1, I1, I2, I3, and R1.
\end{itemize}

Overall, VocBench~3 has been shown to be the most promising for managing multilingual ontologies; however, due to its partial support of OWL~2~DL, it is currently more suited to lightweight ontologies or vocabularies.
For managing an OWL~2~DL ontology, Prot\'eg\'e is the only option available of actively supported tools, yet to implement any of the multilingual options beyond that of annotations is challenging, as the GUI does not provide alternative views of multilingual information.

\section{Discussion}
\label{sec:discussion}

We revisit the questions from Section \ref{sec:introduction} for a brief discussion. 
Concerning RQ1, the available modelling options to develop a multilingual ontology, we described nine principal approaches to do so, which were described in Section~\ref{sec:modelling_multilinguality}. The ad hoc visual language may further help to compare or extend any of the approaches, as well as `allocate' one's ontology into one of the approaches when developing a multilingual ontology, or serve as model for a module that could extend an ODE. It also served in the assessment of the extant ontologies and tooling support for developing multilingual ontologies, of which the main observations are that there is very little tooling support (RQ3) and mostly only simple labels are used (RQ2a), as described in Section~\ref{sec:managing_multilinguality} and Section~\ref{sec:review_ontologies_bioportal}, respectively. 
For the tooling, in particular the interaction with the ontology (localisation and verbalisation) and support for OWL DL or OWL 2 DL can be improved upon, as well as role and versioning support since a multilingual ontology is expected to be a collaboration between multiple parties simultaneously.

Percentages of permeation of multilingual ontologies (RQ2b-d) showed that there is a low uptake with 6.77\% for the BioPortal dataset and 15.74\% in the LOV dataset. Even for the higher LOV percentage, the ontologies are substantially smaller than in BioPortal and they are mostly thesauri and vocabularies, i.e., lightweight ontologies at best. In time, although the 433 multilingual ontologies from Watson (1.7\% of 25~467) is higher than the number found here, this may indeed be less or, perhaps more likely, due to sampling and inclusion criteria, since the ratio of multilingual ontologies has increased. For instance, only `production' ontologies from BioPortal and vocabularies returning a `200' HTTP status code in LOV were considered, but if ontologies with alternative statuses were also analysed, there may be more, since prior versions may also be multilingual and new alpha or beta versions might have made use of an alternative modelling option.  

Overall, the amount of research on multilingual ontologies especially in the 2010s has not translated to a widespread, or even a modest, uptake of the development of multilingual ontologies or tooling support for their development. The data, be it in RDF or similar formats, are distinctly multilingual and increasingly so (recall Section~\ref{sec:related_work} with, among others, \cite{Ell:Vrandecic:Simperl:2011,Kaffee:Simperl:2018}, and Figure~\ref{fig:timeline}), however, and data needs models for management, for which ontologies are exceedingly suited. There may be many reasons for these two mismatches, which could provide ample options for future work. 
Among others, one could conduct a survey with ontology engineers to attempt to elucidate the reasons for low multilinguality uptake or try to estimate how often the `simple' labelling options do not suffice to be needing a more extensive linguistic model.

\section{Conclusion}
\label{sec:conclusion}

The review of multilingual ontologies showed that there is comparatively limited uptake, and even then with only multilingual labels and a limited number of languages. In this assessment, we devised an ad hoc visualisation method to facilitate comparison of the 9 identified mechanisms of modelling multilinguality in (the TBox of) ontologies, and created a set of tool requirements for managing multilingual ontologies. A review of seven relevant ontology editors showed that there are significant gaps in tooling support, with no single tool that supports all the identified requirements, but with VocBench 3 arriving the closest. Multilingualism in production-level BioPortal ontologies revealed that English was used in every multilingual ontology, which  was followed by French and German, with English, Italian and French for LOV, reaching 6.77\% for the BioPortal dataset and 15.74\% in the LOV dataset overall. 

Regarding future research, it may first need to be established why there is limited uptake of multilingualism. 
A comparison of the 9 principal approaches `in action', be it for ontology development or a particular ontology-driven task, may shed light on feasibility of the modelling approaches and of the tooling support and shortcomings.
Surveys among ontology developers may further elucidate why multilingual ontologies were developed, or explicitly not pursued.

\subsubsection*{Acknowledgments} This work was financially supported by Hasso Plattner Institute for Digital Engineering through the HPI Research School at UCT [FGW] and the National Research Foundation (NRF) of South Africa (Grant Number 120852) [CMK].

\bibliography{bibliography}

\end{document}